\documentclass[runningheads]{llncs}

\usepackage{eccv}

\usepackage{eccvabbrv}

\usepackage{graphicx}
\usepackage{booktabs}

\usepackage[accsupp]{axessibility}  % Improves PDF readability for those with disabilities.

\usepackage{hyperref}

\usepackage{orcidlink}
\usepackage{multirow}
\usepackage{colortbl}
\usepackage{pifont}
\usepackage{xcolor}
\usepackage{wrapfig}

\DeclareMathOperator*{\argmin}{arg\,min}

\begin{document}

\title{Modality-Aware Out-of-Distribution Detection for Multi-Modal Action Recognition}

\titlerunning{Modality-Aware Out-of-Distribution Detection}

% Second Author\inst{2,3}\orcidlink{1111-2222-3333-4444} \and
% Third Author\inst{3}\orcidlink{2222--3333-4444-5555}}

\author{Lars Doorenbos\inst{1,2}\orcidlink{0000-0002-0231-9950} \and
Duc Manh Vu\inst{1}\orcidlink{0009-0004-2199-0389} \and
Serdar Ozsoy\inst{1}\orcidlink{0000-0003-1765-4708} \and
Juergen Gall\inst{1,2}\orcidlink{0000-0002-9447-3399}}

\authorrunning{L.~Doorenbos et al.}

\institute{University of Bonn, Germany \and
Lamarr Institute for Machine Learning and Artificial Intelligence, Germany\\
\email{\{doorenbos,soezsoy,gall\}@iai.uni-bonn.de}}

\maketitle

%!TEX root = main.tex

\newif\ifdraft
% \draftfalse
\drafttrue

\definecolor{orange}{rgb}{1,0.5,0}
\definecolor{gr}{rgb}{0,0.65,0}
\definecolor{mygray}{gray}{0.95}

\ifdraft
 \newcommand{\RS}[1]{{\color{red}{\bf RS: #1}}}
 \newcommand{\rs}[1]{{\color{red}#1}}
 \newcommand{\PMN}[1]{{\color{orange}{\bf PMN: #1}}}
 \newcommand{\pmn}[1]{{\color{orange}#1}}
 \newcommand{\LD}[1]{{\color{blue}{\bf LD: #1}}}
 \newcommand{\ld}[1]{{\color{blue}#1}}
 \newcommand{\old}[1]{{\color{gr}#1}}
\else
 \renewcommand{\sout}[1]{}
 \newcommand{\RS}[1]{{\color{red}{}}}
 \newcommand{\rs}[1]{#1}
 \newcommand{\PMN}[1]{{\color{red}{}}}
 \newcommand{\pmn}[1]{#1}
\fi

\newcommand{\framework}{\textsc{\ld{TODO}}\xspace}
\definecolor{lgray}{gray}{0.9}
\newcommand{\real}{\mathbb{R}}
\newcommand{\x}{\mathbf{x}}
\newcommand{\z}{\mathbf{z}}
\newcommand{\y}{\mathbf{y}}
\newcommand{\haty}{\hat{\y}}
\newcommand{\w}{\mathbf{w}}
\renewcommand{\d}{\mathbf{d}}
\newcommand{\D}{\mathcal{D}}
\newcommand{\X}{\mathcal{X}}
\newcommand{\Z}{\mathcal{Z}}
\newcommand{\J}{\mathbf{J}}
\newcommand{\bZ}{\mathbf{Z}}
\newcommand{\M}{\mathcal{M}}
\newcommand{\I}{\mathcal{I}}
\newcommand{\jacobian}{\mathbf{J}}
\newcommand{\balpha}{\bm{\alpha}}
\newcommand{\pkde}{p_{\textrm{kde}}}
\newcommand{\psv}{p_{\balpha}}
\newcommand{\f}{\mathbf{f}}
\newcommand{\g}{\mathbf{g}}
\newcommand{\F}{\mathcal{F}}
\renewcommand{\a}{\mathbf{a}}

\newcommand{\MSCL}{{\bf{MSCL}}}
\newcommand{\PANDA}{{\bf{PANDA}}}
\newcommand{\MKD}{{\bf{MKD}}}
\newcommand{\SSD}{{\bf{SSD}}}
\newcommand{\DNtwo}{{\bf{DN2}}}
\newcommand{\MHRot}{{\bf{MHRot}}}
\newcommand{\DDV}{{\bf{DDV}}}
\newcommand{\IC}{{\bf{IC}}}
\newcommand{\HierAD}{{\bf{HierAD}}}
\newcommand{\Glow}{{\bf{Glow}}}
\newcommand{\MahaAD}{{\bf{MahaAD}}}
\newcommand{\CFlow}{{\bf{CFlow}}}
\newcommand{\NL}{{\bf{NL-Invs}}}
\newcommand{\DIF}{{\bf{DIF}}}

\newcommand{\uniclass}{\emph{uni-class}}
\newcommand{\uniano}{\emph{uni-ano}}
\newcommand{\unimed}{\emph{uni-med}}
\newcommand{\shiftlowres}{\emph{shift-low-res}}
\newcommand{\shifthighres}{\emph{shift-high-res}}

\newcommand{\thyroid}{\emph{thyroid}}
\newcommand{\bc}{\emph{breast cancer}}
\newcommand{\speech}{\emph{speech}}
\newcommand{\pg}{\emph{pen global}}
\newcommand{\shuttle}{\emph{shuttle}}
\newcommand{\kdd}{\emph{KDD99}}

\newcommand{\guood}{{\bf{General U-OOD}}}
\newcommand{\shd}{{\bf{Shallow U-OOD}}}

\newcommand{\xmark}{\ding{55}}%
\newcommand{\cmark}{\ding{51}}%

\begin{abstract}

The incorporation of additional modalities into action recognition models increases their performance across a wide range of settings. However, how this additional information can contribute to making the models more robust remains underexplored, particularly for the case of multi-modal out-of-distribution (OOD) detection. 
While methods exist that regularize the multi-modal training process with OOD detection in mind, they still apply off-the-shelf OOD detectors designed for the uni-modal case during inference, discarding important information. 
Based on an interesting relationship we find between the multi-modal and uni-modal predictions, we propose to use this signal to build a post-hoc detector explicitly designed for the multi-modal scenario. 
We combine this new source of information with a feature-space score, which detects off-manifold samples in the multi-modal space, and normalize them by the multi-modal logits. 
In doing so, the proposed hybrid detector is compatible with existing training-time approaches and consistently improves performance.
Experiments on a wide range of established datasets from the MultiOOD benchmark show that, on average, our approach outperforms the state of the art. 
Our results show the importance of explicitly considering the different modalities at inference time for multi-modal OOD detection.

\end{abstract}    
\section{Introduction}
\label{sec:intro}

Integrating additional modalities into action recognition systems offers a powerful means to enhance their capabilities. The complementary information present in, for example, audio and optical flow enables these systems to consider aspects impossible to obtain from RGB frames alone, greatly increasing their success~\cite{carreira2017quo,cheron2015p,shaikh2024multimodal,sun2022human}. Yet, despite the large interest in improving their performance, little attention has been given to how robust these systems remain when operating outside of controlled settings.

One of the key challenges in designing robust action recognition models, as well as machine learning systems more broadly, is identifying when test samples differ from those observed during training. These \textit{out-of-distribution} (OOD) samples will be confidently handled as belonging to one of the training classes, resulting in silent failures that negatively impact their reliability. To combat this, OOD detection~\cite{hendrycks17msp,yang2024generalized} methods learn scoring functions that measure the level of \textit{out-of-distributionness} of test samples, such that they can be filtered out. However, most current methods are designed for the uni-modal case, leaving multi-modal classifiers without tailored OOD detectors. 

\begin{figure}[t]
  \centering
  \setlength\tabcolsep{12.5pt}
  \begin{tabular}{cc}
    \includegraphics[width=0.4\linewidth]{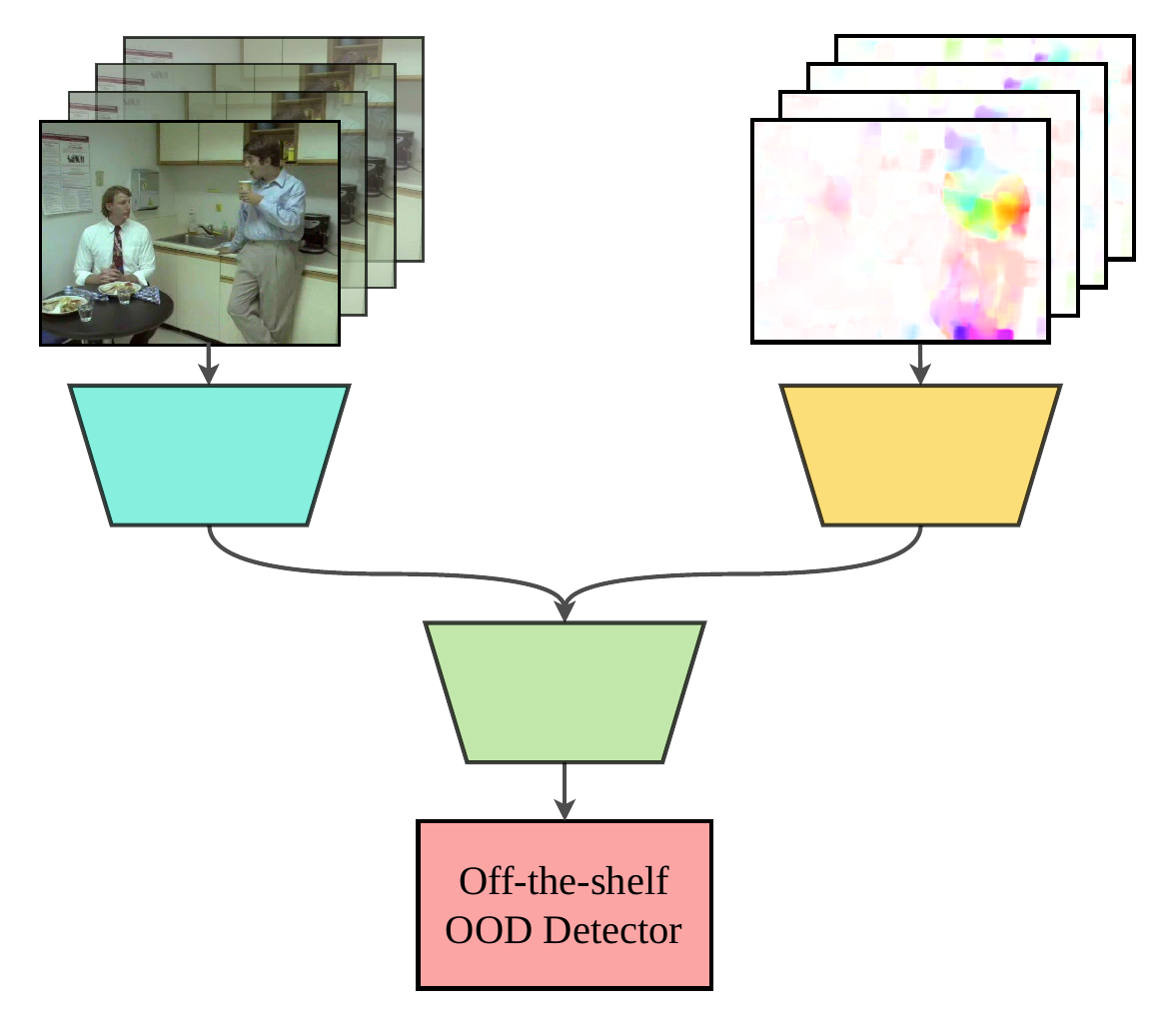} &
    \includegraphics[width=0.4\linewidth]{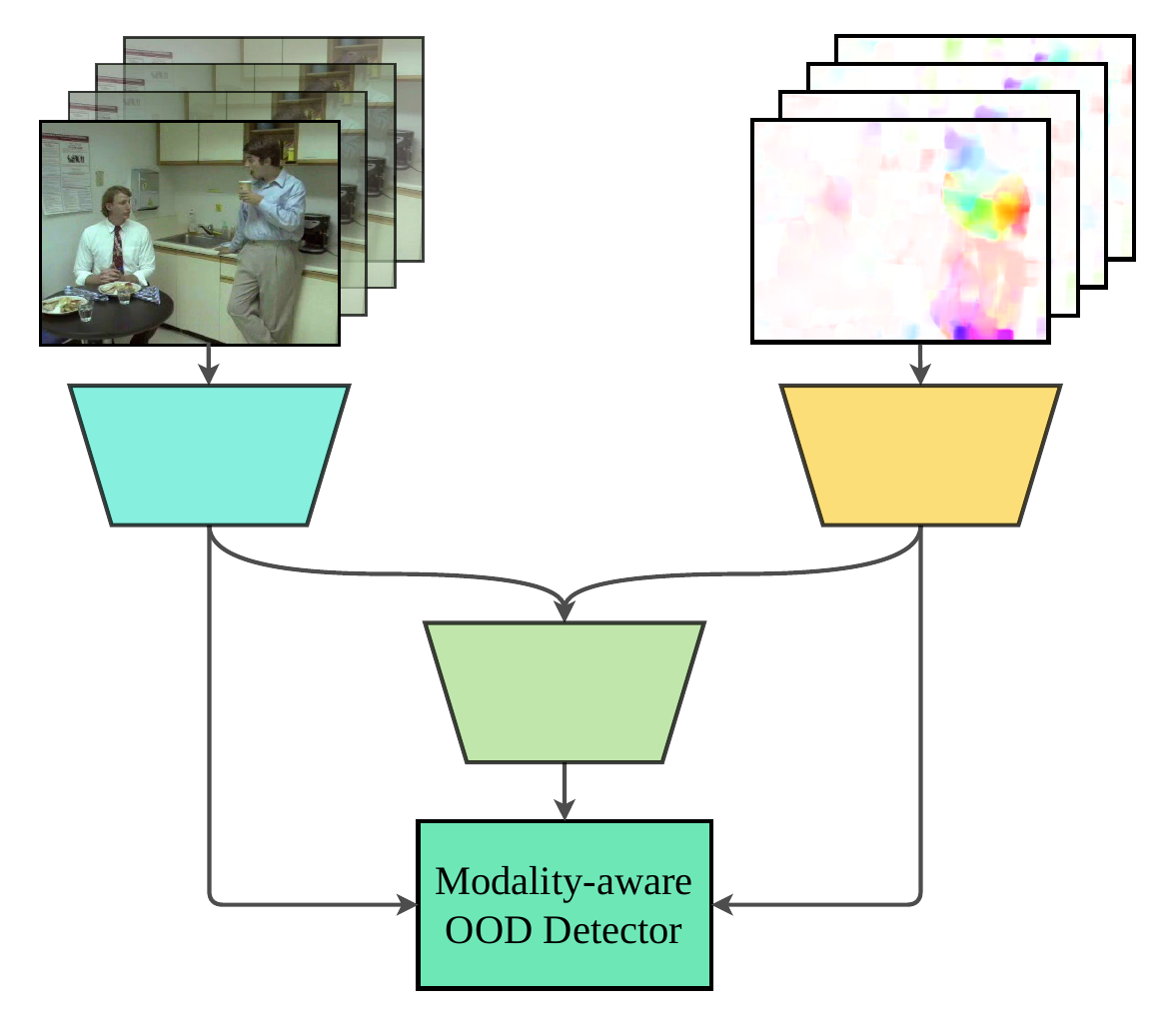} \\
    (a) & (b) \\
  \end{tabular}
    \caption{\textbf{Motivation.} a) Current methods for multi-modal OOD detection apply off-the-shelf detectors designed for the uni-modal case at inference time. b) In contrast, we propose a modality-aware OOD detector that improves multi-modal OOD detection.}
    \label{fig:teaser}
\end{figure}

Only recently, a handful of works have started to consider the multi-modal video OOD detection problem~\cite{dong2024multiood,li2025dpu,liu2025extremely}, following the introduction of the first multi-modal OOD detection benchmark~\cite{dong2024multiood}. They do so by introducing training-time regularization methods that try to enhance the OOD detection performance of post-hoc detectors at inference time. 
However, they do not propose new OOD detectors that leverage multiple modalities at inference time. Instead, they apply off-the-shelf \textit{uni-modal} detectors to the joint representation of the modalities, as illustrated in Fig.~\ref{fig:teaser}. In this way, the multi-modality is not fully exploited for OOD detection at inference time. 

In this work, we address this gap and propose a post-hoc \textit{modality-aware} OOD detector that is motivated by an interesting relationship we identify between the uni-modal and multi-modal predictions for action recognition: there is a high correlation between the multi-modal prediction and an aggregation of the uni-modal predictions for ID samples. 
In contrast, when presented with OOD samples, this relation between uni- and multi-modal predictions will be altered. As such, we find that the relations between the single- and multi-modal decisions are a strong indication of normality. Importantly, this property separates in-distribution (ID) and OOD samples from a different perspective than the standard feature- or prediction-based methods by modeling the impact of multi-modal integration. Indeed, we show that the relation between uni-modal and multi-modal predictions provides a strong cue for OOD samples ignored by previous methods.     

Building on these findings, we propose a modality-aware detector that improves OOD detection performance in multi-modal action recognition scenarios. We do so by integrating this information missed by modality-agnostic scoring functions into a hybrid detector through combining the multi-modal score with a feature-based one and normalizing them using the multi-modal logits. By considering these complementary aspects together, our method is able to detect a wide variety of OOD samples.
We evaluate our approach on the Multi-OOD benchmark~\cite{dong2024multiood}, which comprises five datasets, up to three modalities, and a total of 14 experiments. Our approach consistently outperforms previous baselines. On the challenging Kinetics-600 Far-OOD experiments, which have the largest ID variety, we outperform the state of the art by $6.2$ FPR, an improvement of $13.5\%$. 
In summary, our main contributions are:
\begin{itemize}
    \item We identify an interesting relationship between uni-modal and multi-modal predictions that can be used to separate ID from OOD samples.
    \item We propose a modality-aware OOD detector, which incorporates both uni- and multi-modal information into a hybrid detector to make its predictions during inference.
    \item We demonstrate that our method outperforms previous methods across a wide range of experiments, establishing a new state of the art on the Multi-OOD benchmark.
\end{itemize}

\section{Related Work}

\textbf{Uni-modal OOD Detection.} 
Research into uni-modal OOD detection can be broadly categorized into two types. One direction of research involves applying regularization at training time to boost downstream OOD detection performance. This can be done, for instance, by modifying the loss function to encourage calibrated predictions~\cite{devries2018learning,lee2017training}, designing auxiliary self-supervised objectives~\cite{ahmed2020detecting,winkens2020contrastive}, or knowledge distillation~\cite{tang2025simplification,yangstrengthen}. A more recent trend is to explicitly use outliers during training. These outliers can come from large auxiliary datasets~\cite{hendrycks2018deep} or be derived from the training data itself~\cite{doorenbos2024non,du2024dream,du2022towards}. 
During inference, these methods typically rely on scoring functions related to the regularization approach to detect OOD samples, such as the prediction confidence or entropy.
While these methods can be successful, the requirement to train models from scratch limits their general applicability.

Instead, most detectors opt for a post-hoc approach that can be used with an already trained model, as these models natively contain a variety of signals that can detect OOD samples. For instance, the scores can be derived from the predicted probabilities~\cite{hendrycks17msp,liang2018enhancing} or logits~\cite{hendrycks2019scaling,liang2025revisiting}, as the predictions for OOD samples will be lower than those for ID samples. Similarly, back-propagating the gradients for specifically designed loss functions can highlight OOD samples~\cite{behpour2023gradorth,huang2021importance,sharifi2024gradient}, and they will lie outside of the distribution of the training features in the feature space of the trained network~\cite{kamoi2020mahalanobis,lee2018mahalanobis,sun2022out,mueller2025mahalanobis++}.

More recently, there has been a shift toward post-hoc methods that adopt a hybrid approach: rather than considering these signals independently, they combine information from multiple sources. Due to the complementary nature of these different aspects, this leads to improved performance. Examples are ViM~\cite{wang2022vim} and CADRef~\cite{ling2025cadref}, which combine information from both feature space and the logits, and the hybrid versions of GEN~\cite{liu2023gen}, which combine features with probabilities.
In this work, we also adopt a hybrid, post-hoc approach, but introduce and integrate a new source of information unique to the multi-modal case: the relationship between uni-modal and multi-modal predictions.

\textbf{Multi-modal Video OOD Detection.} 
In contrast to the uni-modal case, research into multi-modal OOD detection has only looked at training time regularization until now, without developing new methods to detect OOD samples during inference.
Specifically, the emerging area of multi-modal OOD detection for action recognition started with the introduction of the Multi-OOD benchmark~\cite{dong2024multiood}, along with the Agree-to-Disagree (A2D) algorithm, which serves as the foundation for all subsequent work. A2D is based on the observation that uni-modal predictions tend to agree more for ID samples than for OOD samples.
To further encourage this during training where no OOD data are available, A2D introduces a regularization term to increase the discrepancy between uni-modal predictions after excluding the ground-truth class.
% , thereby encouraging disagreement across the remaining classes. 
Follow-up work improved upon A2D by dynamic weighting~\cite{li2025dpu} and refined outlier synthesis~\cite{liu2025extremely}. At inference time, all these methods apply existing OOD detectors designed for the \textit{uni-modal} case to the trained networks.

Our work differs in an important aspect: we do not introduce a training-time regularization method; instead, we present the first OOD detector that explicitly considers the different modalities at inference time. As a post-hoc method, our approach is training-free and can be used together with any of the previous training-time methods to boost performance. 

%\section{Method}
\section{Modality-Aware OOD detection}

The goal of OOD detection is to determine whether input samples belong to one of the classes seen at training time, in which case they are considered ID, and OOD otherwise. In the case of multi-modal action recognition, these samples consist of observations of the same action in multiple modalities, such as video, audio, and optical flow. 
Formally, for a multi-modal training dataset $\{\x_i, y_i\}_{i=1}^N$ with labels $y \in \mathcal{Y}$, each sample $\x_i \in \mathcal{X}$ consists of $M$ modalities, $\x_i = \{\x_i^{(m)}|m=1,...,M\}$. 
Based on the training dataset, multi-modal OOD detection aims to find a function $g(\x):\mathcal{X} \to \mathbb{R}$ that assigns the degree of \textit{out-of-distributionness} to a multi-modal test sample $\x$ at inference time.

All previous works in multi-modal OOD detection rely on the A2D method~\cite{dong2024multiood,li2025dpu,liu2025extremely}.
While these methods show that the relations between modalities can contain important information for detecting anomalies, they all rely on detectors designed for the uni-modal case to make their decisions at inference time.
This observation raises the question of why these relations are only used during training. While we show the benefits of a scoring function that explicitly uses multi-modality to detect OOD samples at inference time compared to previous approaches in Sec.~\ref{sec:exp}, we first provide an analysis of uni-modal and multi-modal predictions, which motivates our proposed approach. 

To this end, we formalize the typical architecture used in multi-modal OOD detection for action recognition.
We assume access to a multi-modal classification head that combines all modalities, and uni-modal classification heads for each individual modality. This setup is implicitly present in all current multi-modal OOD detection methods, including A2D~\cite{dong2024multiood} and DPU~\cite{li2025dpu}, and it is not a limitation in practice since heads can be easily added to existing multi-modal architectures, as we show in the supplementary material. At its core, the network is trained with $M+1$ cross-entropy losses. Each modality encoder $e_m$ has a corresponding uni-modal classification head $h_m$, and the multi-modal head $h_f$ operates on all features combined. The loss is then given by:
\begin{equation}
    \mathcal{L}_{cls}(\x, y) = \textrm{CE}(\z^{(f)}, y) + \sum_{m=1}^M \textrm{CE}(\z^{(m)}, y),
\end{equation}
where $\z^{(m)} = h_m \circ e_m(\x^{(m)})$ are the predicted logits for modality $m$, and $\z^{(f)} = h_f(e_1(\x^{(1)}), \cdots, e_M(\x^{(M)}))$ operates on the combined embeddings. 

\begin{figure}[t]
  \centering
  \setlength\tabcolsep{12.5pt}
  \begin{tabular}{cc}
    \includegraphics[width=0.4\linewidth]{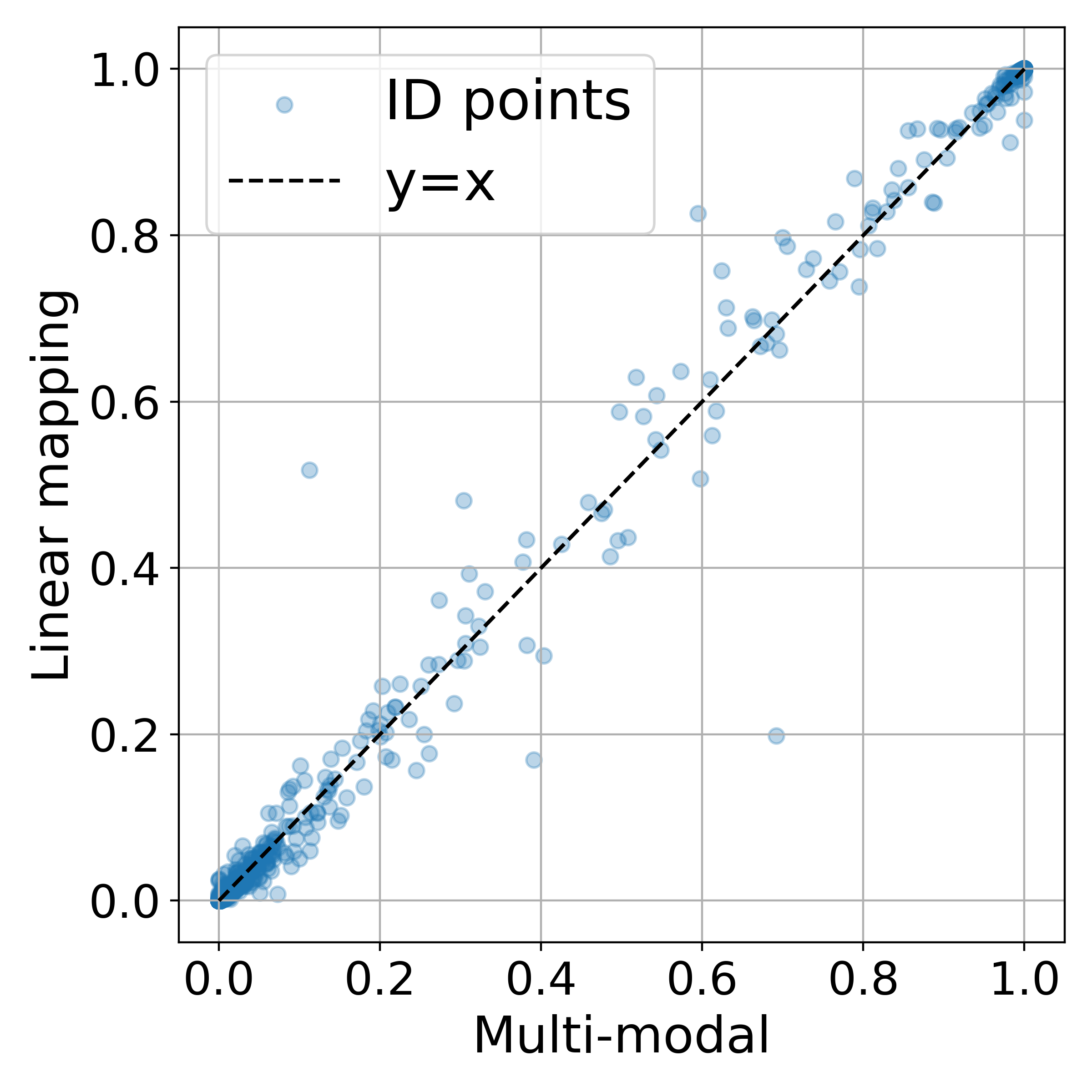} &
    \includegraphics[width=0.4\linewidth]{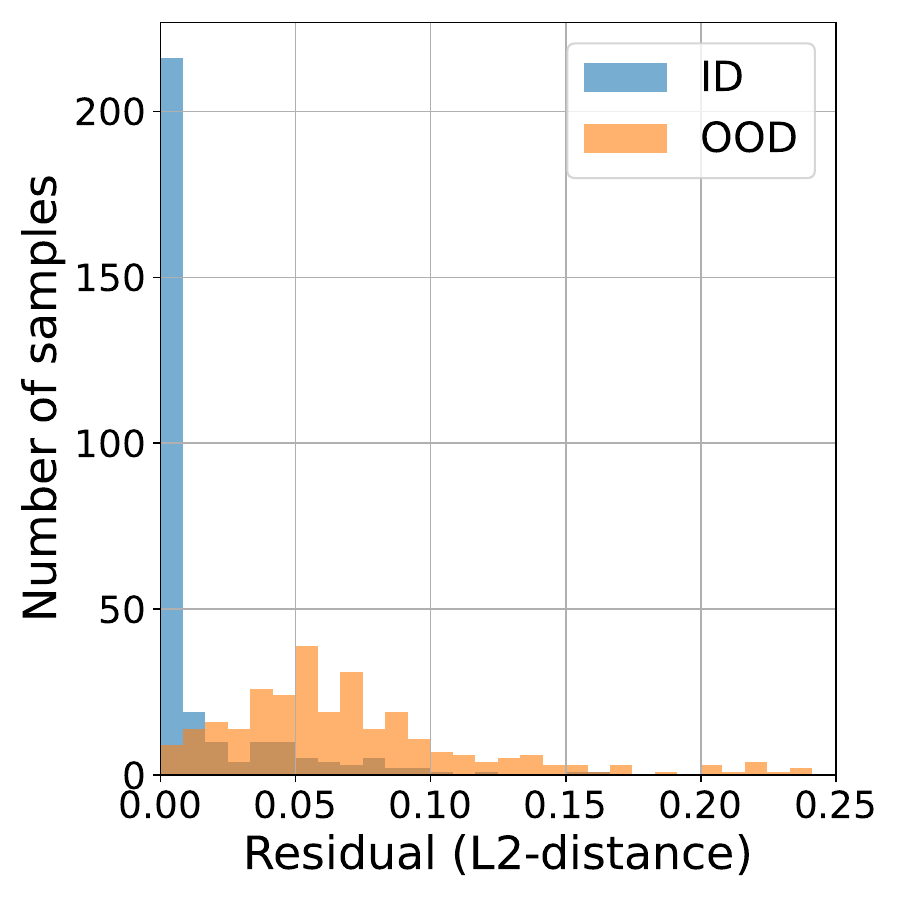} \\
    (a) & (b) \\
  \end{tabular}
    \caption{\textbf{Example of our observed relation between uni- and multi-modal predictions.} (a) The multi-modal predictions can be closely approximated by a linear mapping from the single-modal predictions (Pearson correlation = $0.998$) for ID samples, shown here for HMDB. (b) The distance between the probabilities obtained by the linear mapping from the uni-modal heads versus those of the multi-modal head provides a strong signal for OOD detection.
    }
    \label{fig:motivation}
\end{figure}

Given this structure, we next examine how the uni-modal and multi-modal predictions relate to each other.
Fig.~\ref{fig:motivation} shows an interesting relation between the predictions of the uni-modal heads and the prediction of the multi-modal head that we will exploit. For ID samples, there is a high correlation between the predictions of the multi-modal classification head and a linear mapping from the predictions of the uni-modal heads, whereas this relation is altered for OOD samples, as shown in Fig.~\ref{fig:motivation}(b). 
Building on this intuition, we find that the relations between the single- and multi-modal predictions are a strong indication of normality.

We now describe how we design a novel OOD scoring function that uses this property to detect OOD samples in Section~\ref{sec:score}. Then, we describe in Sections~\ref{sec:integ} and~\ref{sec:final} how we integrate this into a new post-hoc multi-modal OOD detector.

\subsection{Modality-Aware OOD Scoring Function}
\label{sec:score}

To find the linear mapping between the logits predicted by the single-modal heads and the multi-modal head of the trained model, we solve a linear model with intercept on a validation set of $N$ ID samples for each class $c$ separately,
% i.e., without cross-class interactions,
\begin{equation}
\label{eq:linear}
    % \mathbf{w}_c = \min_{\mathbf{w}_c} ||\mathbf{A}_c \mathbf{w}_c - \mathbf{b}_c||_2,
    \tilde{\mathbf{w}}_c = \argmin_{\tilde{\mathbf{w}}_c} ||\tilde{\mathbf{A}}_c \tilde{\mathbf{w}}_c - \mathbf{b}_c||_2,
\end{equation}
where $\tilde{\mathbf{A}}_c = [\mathbf{A}_c, \mathbf{1}] \in \mathbb{R}^{N\times (M+1)}$ contains the uni-modal logits for class $c$ and $\mathbf{b}_c \in \mathbb{R}^{N}$ are the multi-modal logits. Solving this system for $\tilde{\mathbf{w}}_c = [\mathbf{w}_c, w_{0,c}] \in \mathbb{R}^{M+1}$ gives the relative contributions of the modalities and intercept $w_{0,c}$ when approximating the multi-modal logits for that class. 

When applied to a test sample, we use the learned mappings to transform the uni-modal logits to an estimator of the logits of the multi-modal head,
\begin{equation}
    \bar{\z}^{(f)} = \left[\tilde{\mathbf{z}}_1^T\tilde{\mathbf{w}}_1, \cdots,\tilde{\mathbf{z}}_C^T\tilde{\mathbf{w}}_C\right],
\end{equation}
where $\tilde{\mathbf{z}}_c = [\z_c, 1]$ are the logits per modality for class $c$.

After taking the softmax, the error between the predicted and observed multi-modal distribution will be larger for OOD samples, which can be used as a scoring function:
\begin{equation}
\label{eq:dist}
    s(\x) = d(\text{softmax}(\bar{\z}^{(f)}), \text{softmax}(\z^{(f)})),
\end{equation}
with $d(\cdot, \cdot)$ any distance function defined over probability distributions, for which we use the L\textsubscript{2} norm in our experiments.
Conceptually, this implies that the linear mapping between uni-modal and multi-modal predictions will be altered when dealing with abnormal samples, which is shown in Fig.~\ref{fig:motivation}(b). 
In particular, our approach is motivated by the fact that uni-modal representations converge to a shared representation space~\cite{huh2024position}. It is thus intuitive that multi-modal predictions of ID samples can be estimated from uni-modal predictions, as they are in the same vector space. In this work, we find that this shared space does not hold for OOD samples and exploit this for OOD detection.

\subsection{Feature Space Integration}
\label{sec:integ}

The most successful OOD detection algorithms incorporate information from multiple sources, typically combining both features and predicted probabilities (e.g., \cite{wang2022vim}). The relation between uni- and multi-modal predictions, described in the previous section, can add another dimension to these hybrid methods. We now describe how we obtain information from the multi-modal feature space and integrate the predictions to obtain our final detector.

In feature space, cross-modal connections can be modeled by considering the manifold of the concatenated features.
Then, off-manifold directions are strong indicators of OOD samples.
We represent these directions by the principal components with the lowest variance, i.e., the linear invariants~\cite{doorenbos2022data}. Any deviations in these directions constitute off-manifold samples that are OOD.

To obtain the invariants, we perform an Eigen decomposition of the sample covariance matrix $\mathbf \Sigma$ of the normalized concatenated features of dimensionality $D$,
% $\left[e_1(\x^{(1)}), \cdots, e_M(\x^{(M)})\right]$
\begin{equation}
    \mathbf{\Sigma} = \mathbf{Q} \mathbf{\Lambda} \mathbf{Q}^{\intercal},
\end{equation}
where $\mathbf{Q}\in\mathbb{R}^{D \times D}$ contains the Eigenvectors of $\mathbf{\Sigma}$, and $\mathbf{\Lambda} \in \mathbb{R}^{D \times D}$ its Eigenvalues. To address high-dimensional and low-data regimes, we employ the hyper-parameter-free Ledoit-Wolf shrinkage method~\cite{ledoit2004well} to compute $\mathbf{\Sigma}$. Then, in order to focus on the common features of the entire dataset, we use the $k$ Eigenvectors with the smallest Eigenvalues, denoted as $\mathbf Q_{k}\in \mathbb{R}^{k \times D}$ and $\mathbf{\Lambda}_{k}\in \mathbb{R}^{k \times k}$. 
We can set $k$ adaptively by tying it to the number of principal components of the data that jointly explain less than $p$\% of the variance, where $p$ is a hyperparameter of our method. This way, if the training data lies on a lower-dimensional manifold, more dimensions will be used to detect OOD.

\begin{wrapfigure}{r}{5.2cm}
\includegraphics[width=5.2cm]{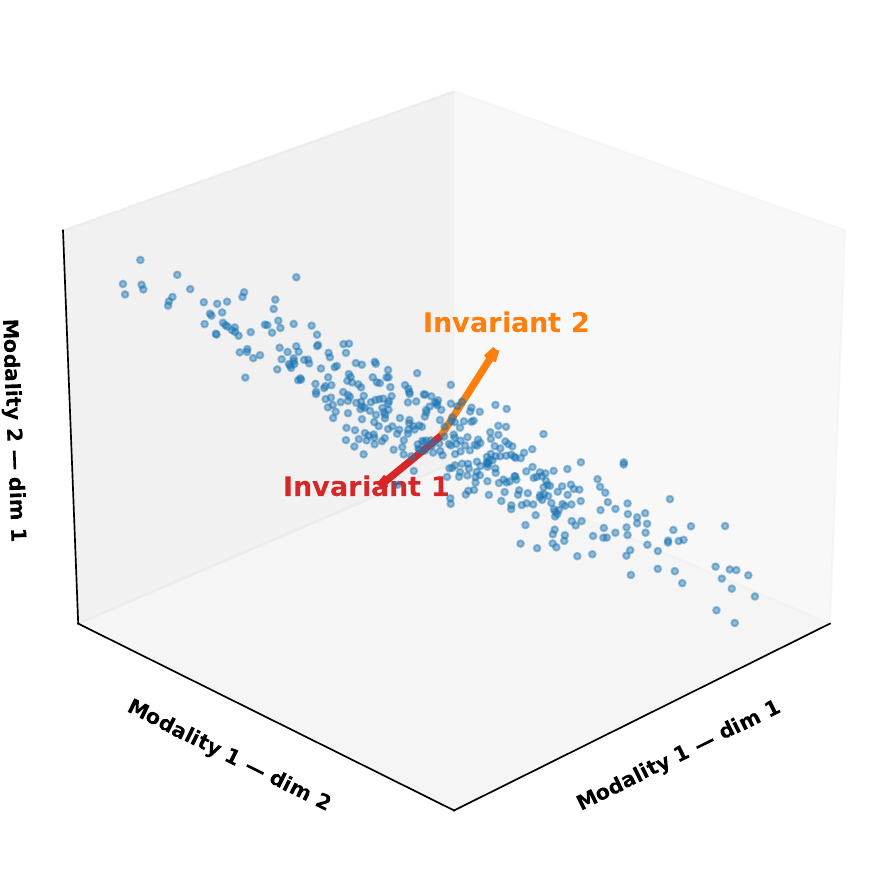}
\caption{\textbf{Finding the invariants on a toy example.} The principal components with the lowest variance describe off-manifold directions in the multi-modal feature space. We use these to detect feature-level OOD samples.}
    \label{fig:feature}
\end{wrapfigure} 

Any deviation in these tight dimensions is a strong signal that a sample lies off-manifold and is OOD. Furthermore, deviations in dimensions with smaller variance should have a larger impact than similar-sized deviations in dimensions with larger variance. Therefore, we use the Mahalanobis distance to compute our feature-level score,
% , rather than relying only on the L2-distance in this lower-dimensional representation as done in~\cite{wang2022vim}
\begin{equation}
    r(\x) = \sqrt{(\x-\mu)^\intercal \mathbf Q_{k}\mathbf \Lambda_{k}^{-1}\mathbf Q_{k}^\intercal(\x-\mu)},
\end{equation}
where $\mu$ is the mean of the training features.

Applying $r(\x)$ to the concatenated features $(e_1(\x^{(1)}), \cdots, e_M(\x^{(M)}))$ of all modalities will find the off-manifold directions in the training set and score test samples accordingly. We show this for a toy example in Fig.~\ref{fig:feature}.

\subsection{Final Score}
\label{sec:final}

Beyond the relation between uni- and multi-modal predictions and feature space, 
the final predictions of the classifier are an indicator of OOD from a third point of view,
and we find that performance can be further improved by taking a holistic approach and combining the three different aspects. 
We incorporate the previous scores with the predicted distribution as virtual logits~\cite{wang2022vim} with an additional scaling factor. We first compute the mean and standard deviation of the highest predicted logit over the validation samples and use it to scale $s(\x)$ and $r(\x)$ to have the same statistics, i.e., 
\begin{align}    
\label{eq:scaling}
    \alpha =  \frac{\sigma_l}{\sigma_s}, \quad \beta = \mu_l - \alpha \mu_s\\
    s'(\x) = \gamma_s (\alpha s(\x) + \beta),
\end{align}
where $\mu_l$, $\sigma_l$, $\mu_s$, and $\sigma_s$ are the means and standard deviations of the highest logit and multi-OOD score, respectively, and analogously construct $r'(\x)$ from $r(\x)$. The hyperparameters $\gamma_s$ and $\gamma_r$ control the relative strength of the components.
Then, the scaled scores are integrated with the predicted logits to obtain the final OOD score,
\begin{equation}
\label{eq:logit}
    g(\x) = \frac{e^{s'(\x)} + e^{r'(\x)}}{\sum_{i=1}^C e^{\z^{(f)}_i} + e^{s'(\x)}+ e^{r'(\x)}}.
\end{equation}

Intuitively, this normalized score acts as a logit for the OOD class: confident predictions into one of the training classes will decrease the OOD probability, whereas abnormal multi-modal representations will increase it. 

\begin{table*}[t]
\caption{\textbf{Kinetics-600 Far-OOD Detection results with video and optical flow.} A2D results taken from~\cite{dong2024multiood}, FM results taken from~\cite{liu2025extremely}. We report the mean and standard deviation over three runs. Our method achieves the best overall results.}
\label{tab:kin_far}
% \vspace{-0.1in}
\centering
% \renewcommand{\arraystretch}{1.00}
% \setlength{\tabcolsep}{7pt} 
% \fontsize{8}{10}\selectfont 
\resizebox{0.9\textwidth}{!}{
\begin{tabular}{r|cccccccc|cc|c}
\toprule %\toprule
\multicolumn{1}{c|}{\multirow{3}{*}{Method}} & \multicolumn{10}{c|}{OOD Datasets}                                       & \multicolumn{1}{l}{\multirow{3}{*}{ Acc }} \\ \cline{2-11}
\multicolumn{1}{c|}{}                        & \multicolumn{2}{c|}{HMDB51}                                 & \multicolumn{2}{c|}{UCF101}                                     & \multicolumn{2}{c|}{HAC}                                        & \multicolumn{2}{c|}{EPIC-Kitchen}  & \multicolumn{2}{c|}{Average}                      & \multicolumn{1}{l}{}                        \\ \cline{2-11}
\multicolumn{1}{c|}{}                        & \multicolumn{1}{l}{FPR $\downarrow$} & \multicolumn{1}{l|}{AUC $\uparrow$}           & \multicolumn{1}{l}{FPR $\downarrow$} & \multicolumn{1}{l|}{AUC $\uparrow$}          & \multicolumn{1}{l}{FPR $\downarrow$} & \multicolumn{1}{l|}{AUC $\uparrow$}  & \multicolumn{1}{l}{FPR $\downarrow$} & \multicolumn{1}{l|}{AUC $\uparrow$}          & \multicolumn{1}{l}{FPR $\downarrow$} & \multicolumn{1}{l|}{AUC $\uparrow$} & \multicolumn{1}{l}{}                        \\ \midrule
A2D\textsubscript{p} & 61.7 & 80.6 & 57.0 & 78.0 & \textbf{42.0} & \textbf{87.0} & 40.1 & 85.1 & 50.2 & 82.7 & 73.7\\
A2D\textsubscript{f} & 63.7 & 79.1 & 62.2 & 75.3 & 46.5 & 85.2 & 36.1 & 88.7 & 52.1 & 82.1 & 72.5\\
A2D\textsubscript{h} & 63.3 & 74.0 & 66.3 & 74.1 & 53.8 & 81.1 & 34.5 & 87.7 & 54.5 & 79.2 & 73.7\\
FM & 62.9 & 74.3 & 67.7 & 74.4 & 54.9 & 80.6 & 33.5 & 87.6 & 54.8 & 79.2 & 73.7 \\
DPU\textsubscript{p} & 54.0\tiny{$\pm$2.4} & 83.2\tiny{$\pm$0.4} & 52.6\tiny{$\pm$1.5} & 81.8\tiny{$\pm$0.4} & 46.6\tiny{$\pm$2.4} & 84.8\tiny{$\pm$1.2} & 30.5\tiny{$\pm$0.5} & 89.7\tiny{$\pm$0.9} & 45.9 & 84.9 & 75.2\\
DPU\textsubscript{f} & 58.5\tiny{$\pm$6.5} & 84.9\tiny{$\pm$0.7} & 57.7\tiny{$\pm$3.1} & 82.9\tiny{$\pm$1.1} & 52.3\tiny{$\pm$3.5} & 84.6\tiny{$\pm$0.8} & \textbf{24.5}\tiny{$\pm$0.5} & \textbf{91.7}\tiny{$\pm$0.9} & 48.3 & 86.0 & 75.2 \\
DPU\textsubscript{h} & 59.7\tiny{$\pm$6.1} & 79.5\tiny{$\pm$3.2} & 59.2\tiny{$\pm$3.6} & 76.7\tiny{$\pm$3.5} & 48.6\tiny{$\pm$2.3} & 82.6\tiny{$\pm$1.3} & 25.6\tiny{$\pm$0.5} & 90.8\tiny{$\pm$0.8} & 48.3 & 82.4 & 75.2\\
\rowcolor{blue!20!white} Ours & \textbf{49.9}\tiny{$\pm$4.2} & \textbf{88.5}\tiny{$\pm$0.6} & \textbf{37.6}\tiny{$\pm$2.3} & \textbf{90.8}\tiny{$\pm$0.8} & 44.2\tiny{$\pm$2.5} & 86.2\tiny{$\pm$1.0} & 27.1\tiny{$\pm$1.2} & 90.9\tiny{$\pm$1.0} & \textbf{39.7} & \textbf{89.1} & 75.2\\
\bottomrule %\bottomrule
\end{tabular}
    }
% \vspace{-0.15in}
\end{table*}

\section{Experiments}\label{sec:exp}

\textbf{Evaluation protocol.}
We follow the Multi-OOD benchmark~\cite{dong2024multiood} with Near- and Far-OOD experiments using Kinetics-600~\cite{carreira2018short}, HMDB51~\cite{kuehne2011hmdb}, UCF101~\cite{soomro2012ucf101}, EPIC-Kitchen~\cite{damen2018scaling}, and HAC~\cite{dong2023simmmdg}. 
We evaluate our method against the strongest reported versions of previous multi-modal OOD methods: A2D with NP-Mix (A2D)~\cite{dong2024multiood}, DPU~\cite{li2025dpu}, and FM~\cite{liu2025extremely}. As these methods are evaluated using a variety of off-the-shelf detectors, we compare against the best-performing variant of each baseline method across three detector categories: probabilities/logits-based, feature-based, and hybrid versions. We denote these with subscripts \textsubscript{p}, \textsubscript{f}, and \textsubscript{h}, respectively. Thus, A2D\textsubscript{f} denotes the best results of A2D with a feature-based scoring function. We give the exact method used for every number and full dataset details in the supplementary material.
Following standard procedure, we report the False Positive Rate at 95\%, True Positive Rate (FPR), and the Area Under the ROC Curve (AUC). We also report the in-distribution accuracy of the classifiers (Acc).

\textbf{Implementation details.}
All methods are implemented with the same architectures. Similar to previous work~\cite{dong2024multiood,li2025dpu,liu2025extremely}, we use the SlowFast~\cite{feichtenhofer2019slowfast} model pre-trained on Kinetics-400 to encode videos. For optical flow, we also use SlowFast pre-trained on Kinetics-400, with a slow-only pathway. We train the baselines using the official settings provided in their repositories across three seeds to assess the variability in results unless specified otherwise. We apply our method to networks trained with DPU, use the L2-distance for $d(\cdot, \cdot)$, and fix $\gamma_s$ to $0.4$, $\gamma_r$ to $1.25$, and $k$ to $p=0.5$\% in all experiments.\footnote{The code is available at \url{https://github.com/LarsDoorenbos/modality-aware-ood}}

\begin{table*}[t]
\caption{\textbf{HDMB54 Far-OOD Detection results with video and optical flow.} FM results taken from~\cite{liu2025extremely}. We report the mean and standard deviation over three runs. Our method achieves the best overall results.}
\label{tab:hmdb_far}
% \vspace{-0.1in}
\centering
% \renewcommand{\arraystretch}{1.00}
% \setlength{\tabcolsep}{7pt} 
% \fontsize{8}{10}\selectfont 
\resizebox{0.9\textwidth}{!}{
\begin{tabular}{r|cccccccc|cc|c}
\toprule %\toprule
\multicolumn{1}{c|}{\multirow{3}{*}{Method}} & \multicolumn{10}{c|}{OOD Datasets}                                       & \multicolumn{1}{l}{\multirow{3}{*}{ Acc }} \\ \cline{2-11}
\multicolumn{1}{c|}{}                        & \multicolumn{2}{c|}{Kinetics-600}                                 & \multicolumn{2}{c|}{UCF101}                                     & \multicolumn{2}{c|}{HAC}                                        & \multicolumn{2}{c|}{EPIC-Kitchen}         & \multicolumn{2}{c|}{Average}               & \multicolumn{1}{l}{}                        \\ \cline{2-11}
\multicolumn{1}{c|}{}                        & \multicolumn{1}{l}{FPR $\downarrow$} & \multicolumn{1}{l|}{AUC $\uparrow$}           & \multicolumn{1}{l}{FPR $\downarrow$} & \multicolumn{1}{l|}{AUC $\uparrow$}          & \multicolumn{1}{l}{FPR $\downarrow$} & \multicolumn{1}{l|}{AUC $\uparrow$}          & \multicolumn{1}{l}{FPR $\downarrow$} & \multicolumn{1}{l|}{AUC $\uparrow$}  & \multicolumn{1}{l}{FPR $\downarrow$} & \multicolumn{1}{l|}{AUC $\uparrow$} & \multicolumn{1}{l}{}                        \\ \midrule
A2D\textsubscript{p} & 24.1\tiny{$\pm$5.7} & 94.3\tiny{$\pm$1.3} & 35.5\tiny{$\pm$2.7} & 90.2\tiny{$\pm$0.1} & 22.7\tiny{$\pm$4.4} & 94.6\tiny{$\pm$0.9} & 9.5\tiny{$\pm$5.1} & 97.4\tiny{$\pm$1.1} & 23.0 & 94.1 & 86.9 \\
A2D\textsubscript{f} & 11.7\tiny{$\pm$0.4} & 97.1\tiny{$\pm$0.2} & 31.6\tiny{$\pm$2.0} & 89.9\tiny{$\pm$1.0} & 9.4\tiny{$\pm$1.4} & 97.7\tiny{$\pm$0.4} & 10.9\tiny{$\pm$1.5} & 96.5\tiny{$\pm$0.3} & 15.9 & 95.3 & 86.9 \\
A2D\textsubscript{h} & 10.5\tiny{$\pm$1.8} & 97.8\tiny{$\pm$0.3} & 28.6\tiny{$\pm$2.1} & 91.6\tiny{$\pm$0.7}  & 8.0\tiny{$\pm$1.7} & 98.3\tiny{$\pm$0.3} & 5.7\tiny{$\pm$1.4} & 98.4\tiny{$\pm$0.3} & 13.2 & 96.5 & 86.9\\
FM & 19.6 & 94.7 & 34.3 & 90.1 & 16.0 & 95.5 & 10.2 & 96.3 & 20.0 & 94.2 & 87.0 \\
DPU\textsubscript{p} & 19.3\tiny{$\pm$0.2} & 96.0\tiny{$\pm$0.4} & 27.2\tiny{$\pm$1.9} & 93.7\tiny{$\pm$0.5} & 14.8\tiny{$\pm$1.3} & 96.9\tiny{$\pm$0.4} & 6.3\tiny{$\pm$7.0} & \textbf{98.7}\tiny{$\pm$1.2} & 16.9 & 96.3 & 87.4\\
DPU\textsubscript{f} & 10.4\tiny{$\pm$1.7} & 97.7\tiny{$\pm$0.3} & 24.5\tiny{$\pm$1.3} & 92.4\tiny{$\pm$0.7} & 9.2\tiny{$\pm$1.3} & 97.9\tiny{$\pm$0.7} & 10.4\tiny{$\pm$2.0} & 96.8\tiny{$\pm$0.8} & 13.6 & 96.2 & 87.4 \\
DPU\textsubscript{h} & 8.5\tiny{$\pm$1.0} & \textbf{98.1}\tiny{$\pm$0.0} & 23.2\tiny{$\pm$3.6} & 93.0\tiny{$\pm$0.8} & \textbf{6.8}\tiny{$\pm$0.8} & \textbf{98.6}\tiny{$\pm$0.3} & 6.0\tiny{$\pm$1.2} & 98.4\tiny{$\pm$0.5} & 11.1 & \textbf{97.0} & 87.4\\
\rowcolor{blue!20!white} Ours & \textbf{8.3}\tiny{$\pm$0.5} & 97.6\tiny{$\pm$0.5} & \textbf{20.4}\tiny{$\pm$1.3} & \textbf{94.8}\tiny{$\pm$0.8} & 8.2\tiny{$\pm$1.4} & 97.1\tiny{$\pm$0.6} & \textbf{3.6}\tiny{$\pm$1.9} & 98.3\tiny{$\pm$0.8} & \textbf{10.1} & \textbf{97.0} & 87.4\\
\bottomrule %\bottomrule
\end{tabular}
    }
% \vspace{-0.15in}
\end{table*}

\begin{table*}[t]
\caption{\textbf{Multi-modal Near-OOD Detection results using video and optical flow.} We report the mean and standard deviation over three runs. ID accuracies are shown in the supplementary material. Our method reaches the highest average performance. }
\label{tab:near}
% \vspace{-0.1in}
\centering
% \renewcommand{\arraystretch}{1.00}
% \setlength{\tabcolsep}{7pt} 
% \fontsize{8}{10}\selectfont 
\resizebox{0.9\textwidth}{!}{
\begin{tabular}{r|cccccccc|cc}
\toprule %\toprule
\multicolumn{1}{c|}{\multirow{3}{*}{Method}} & \multicolumn{10}{c}{Datasets}                                        \\ \cline{2-11}
\multicolumn{1}{c|}{}                        & \multicolumn{2}{c|}{HMDB51}                                 & \multicolumn{2}{c|}{UCF101}                                     & \multicolumn{2}{c|}{Kinetics-600}                                        & \multicolumn{2}{c|}{EPIC-Kitchen}          & \multicolumn{2}{c}{Average}                                 \\ \cline{2-11}
\multicolumn{1}{c|}{}                        & \multicolumn{1}{l}{FPR $\downarrow$} & \multicolumn{1}{l|}{AUC $\uparrow$}           & \multicolumn{1}{l}{FPR $\downarrow$} & \multicolumn{1}{l|}{AUC $\uparrow$}          & \multicolumn{1}{l}{FPR $\downarrow$} & \multicolumn{1}{l|}{AUC $\uparrow$}     & \multicolumn{1}{l}{FPR $\downarrow$} & \multicolumn{1}{l|}{AUC $\uparrow$}       & \multicolumn{1}{l}{FPR $\downarrow$} & \multicolumn{1}{l}{AUC $\uparrow$}                   \\ \midrule
A2D\textsubscript{p} & 38.4\tiny{$\pm$3.4} & 89.1\tiny{$\pm$0.7} & 8.0\tiny{$\pm$1.8} & 98.3\tiny{$\pm$0.2} & \textbf{61.2}\tiny{$\pm$2.9} & \textbf{77.4}\tiny{$\pm$0.7} & 70.0\tiny{$\pm$4.9} & 71.7\tiny{$\pm$3.4} & 44.4 & 84.1 \\
A2D\textsubscript{f} & 36.5\tiny{$\pm$2.2} & 89.8\tiny{$\pm$0.2} & 12.7\tiny{$\pm$0.4} & 97.4\tiny{$\pm$0.2} & 64.0\tiny{$\pm$0.2} & 77.1\tiny{$\pm$0.8} & 75.8\tiny{$\pm$2.3} & 68.1\tiny{$\pm$1.9} & 47.3 & 83.1 \\
A2D\textsubscript{h} & 36.2\tiny{$\pm$0.8} & 88.2\tiny{$\pm$0.5} & {7.0}\tiny{$\pm$0.9} & \textbf{98.4}\tiny{$\pm$0.2} & 63.2\tiny{$\pm$3.3} & 77.3\tiny{$\pm$1.3} & 71.5\tiny{$\pm$2.8} & 67.1\tiny{$\pm$2.2} & 44.5 & 82.8 \\
DPU\textsubscript{p} & 35.9\tiny{$\pm$2.4} & 89.0\tiny{$\pm$0.4} & 10.0\tiny{$\pm$4.6} & 97.7\tiny{$\pm$0.8} & 63.4\tiny{$\pm$0.2} & 76.9\tiny{$\pm$0.7} & 70.7\tiny{$\pm$2.1} & 68.5\tiny{$\pm$0.7} & 45.0 & 83.0 \\
DPU\textsubscript{f} & 35.4\tiny{$\pm$3.3} & 89.6\tiny{$\pm$0.5} & 10.6\tiny{$\pm$1.8} & 97.6\tiny{$\pm$0.2} & 66.6\tiny{$\pm$1.1} & 76.6\tiny{$\pm$0.8} & 76.0\tiny{$\pm$3.7} & 67.0\tiny{$\pm$1.8} & 47.2 & 82.7 \\
DPU\textsubscript{h} & 34.9\tiny{$\pm$2.2} & 89.1\tiny{$\pm$0.8} & 7.8\tiny{$\pm$0.5} & 98.2\tiny{$\pm$0.2} & 63.8\tiny{$\pm$0.3} & 77.1\tiny{$\pm$0.7} & 72.8\tiny{$\pm$1.8} & 66.5\tiny{$\pm$0.2} & 44.8 & 82.7 \\
\rowcolor{blue!20!white} Ours & \textbf{33.8}\tiny{$\pm$3.0} & \textbf{90.3}\tiny{$\pm$0.3} & \textbf{6.5}\tiny{$\pm$1.6} & \textbf{98.4}\tiny{$\pm$0.3} & 63.0\tiny{$\pm$1.1} & 77.0\tiny{$\pm$0.4} & \textbf{69.0}\tiny{$\pm$1.3} & \textbf{72.1}\tiny{$\pm$0.8} & \textbf{43.1} & \textbf{84.5} \\
\bottomrule %\bottomrule
\end{tabular}
    }
% \vspace{-0.15in}
\end{table*}

\begin{table}[t]
\begin{center}
\caption{\textbf{Triple-modal OOD detection.} \textit{Left:} Results with audio, video, and optical flow over three seeds for EPIC-Kitchens Near-OOD and EPIC-Kitchens:Kinetics Far-OOD. \textit{Right:} Ablating the benefits of multiple modalities for OOD detection on EPIC-Kitchens:Kinetics. Integrating all modalities gives the best results.
}
 \begin{tabular}{c|ccccc}
    \toprule
      & Near & Far & \multicolumn{2}{c}{Average} \\
       & AUC & AUC & FPR & AUC \\
    \midrule
    A2D\textsubscript{p} & \textbf{73.3}\tiny{$\pm$2.6} & 78.8\tiny{$\pm$3.3} & 64.2 & 76.0\\
    A2D\textsubscript{f} & 65.8\tiny{$\pm$2.4} & 87.5\tiny{$\pm$4.0} & 58.7 & 76.6 \\
    A2D\textsubscript{h} &  52.0\tiny{$\pm$5.8} & \textbf{98.0}\tiny{$\pm$0.8} & 45.9 & 75.0\\
  \rowcolor{blue!20!white} Ours & 65.5\tiny{$\pm$2.4} & \textbf{98.0}\tiny{$\pm$0.3} & \textbf{41.6} & \textbf{81.8}\\   
   \hline
\end{tabular}
\quad\quad
 \begin{tabular}{ccc|ccc}
    \toprule
     \multirow{2}{*}{Video} & \multirow{2}{*}{Optical Flow} & \multirow{2}{*}{Audio} & \multicolumn{2}{c}{Far}\\
       & & & FPR & AUC \\
    \midrule
  \cmark & \xmark & \xmark & 15.4 & 97.0\\
  \xmark & \cmark & \xmark & 28.9 & 93.4 \\
  \xmark & \xmark & \cmark & 48.5 & 86.7 \\
  \cmark & \cmark & \xmark & 12.8 & 97.2 \\
  \rowcolor{blue!20!white} \cmark & \cmark & \cmark & \textbf{10.8} & \textbf{98.0} \\   
   \hline
\end{tabular}
\label{tab:modalities}
% }
\end{center}
\end{table}

\subsection{Main Results}

\textbf{Far-OOD.}
We show the results for Far-OOD detection with Kinetics-600 as the in-distribution in Tab.~\ref{tab:kin_far} and with HMDB51 as the in-distribution in Tab.~\ref{tab:hmdb_far}. 
On the Kinetics-600 experiments, we set a new state-of-the-art with HMDB51 and UCF101 as the OOD dataset, and reach the second-best AUC on the remaining two. On average, this leads to an improvement of $6.2$ FPR over the next best method. Importantly, this baseline uses the same trained network, highlighting the benefit of our scoring function.
We find that the same pattern holds when HMDB51 is used as the in-distribution: our method achieves the state-of-the-art with the overall best performance, an average FPR of $10.1$ and AUC of $97.0$.

\textbf{Near-OOD.}
The Near-OOD results in Tab.~\ref{tab:near} show that our approach also works well in this setting. We achieve the best results in all cases except for Kinetics, leading to an overall improvement of $1.3$ in average FPR over the next-best method, A2D\textsubscript{p}. Interestingly, A2D\textsubscript{p} scored far worse on Far-OOD: for instance, its average FPR is $12.9$ points worse with HMDB as the ID (Tab.~\ref{tab:hmdb_far}). For methods using the same network weights, we improve by $1.7$ average FPR over DPU\textsubscript{h}. In general, these baselines are less consistent, with DPU\textsubscript{h} being the best DPU version for Near-OOD, but the worst on Kinetics Far-OOD.

\textbf{Three modalities.}
We show triple-modal results in Tab.~\ref{tab:modalities}(left) where we use A2D to train the backbone, as DPU does not provide training code for the triple-modal case. In this setting, the probability-based method A2D\textsubscript{p} scores very well in the Near-OOD setting, but $19.2$ AUC points behind our method for Far-OOD. For the hybrid method A2D\textsubscript{h} this is reversed, scoring high for Far-OOD but $13.5$ points below our method on Near-OOD. Overall, our method achieves the best average performance. Furthermore, these results also confirm that our method works when A2D is used to train the network.

\subsection{The benefits of multiple modalities}

The complementary information present in different modalities increases the capabilities of action recognition systems. Here, we show that the same holds for our proposed method. We conduct an experiment using different combinations of the three modalities for the EPIC-Kitchens:Kinetics task and present the results in Tab.~\ref{tab:modalities}(right).
We find that video is the strongest modality of the three, followed by optical flow. By itself, audio performs $33.1$ FPR worse than video. However, when integrated with the other modalities in our method, audio still brings an improvement of $2.0$ FPR over video and flow alone.

\subsection{Effectiveness under different fusion mechanisms}
In the previous sections, we followed the late fusion architecture design used in all previous multi-modal OOD studies. Here, we additionally validate our approach on attention-based early- and mid-fusion networks using the same settings as before. To do so, we train two DPU networks replacing the multi-modal head $h_f$ with a transformer: one with cross-attention layers throughout (early-fusion) and one with cross-attention starting halfway through the network (mid-fusion). 

We show the results for both Near- and Far-OOD in Tab.~\ref{tab:fusion}, where we find similar trends as the main results: our method outperforms the baselines in this regime by $5.2$-$5.8$ average FPR. 
Note that the performance of all methods can be generally higher or lower than previous experiments, depending on how well the architecture fits the action recognition task, as reflected in the ID accuracy. Nonetheless, all methods use the same network, so this does not affect the OOD detection performance comparisons in the table.

\begin{table*}[t]
\caption{\textbf{Results with early- and mid-fusion networks} for Kinetics-600 as the ID dataset. Our method is also successful with attention-based early- and mid-fusion.}
\label{tab:fusion}
% \vspace{-0.1in}
\centering
% \renewcommand{\arraystretch}{1.00}
% \setlength{\tabcolsep}{7pt} 
% \fontsize{8}{10}\selectfont 
\resizebox{0.9\textwidth}{!}{
\begin{tabular}{r|ccc|cccccccc|cc|c}
\toprule %\toprule
\multicolumn{1}{c|}{\multirow{3}{*}{Method}} & \multicolumn{3}{c|}{Near-OOD}      & \multicolumn{10}{c|}{Far-OOD Datasets}                                       &  \\ \cline{2-14}
\multicolumn{1}{c|}{}                        & \multicolumn{3}{c|}{Kinetics}       & \multicolumn{2}{c|}{HMDB51}                                 & \multicolumn{2}{c|}{UCF101}                                     & \multicolumn{2}{c|}{HAC}                                        & \multicolumn{2}{c|}{EPIC-Kitchen}  & \multicolumn{2}{c|}{Average}                      & \multicolumn{1}{l}{}                        \\ \cline{2-14}
\multicolumn{1}{c|}{}                      & \multicolumn{1}{l}{FPR $\downarrow$} & \multicolumn{1}{l}{AUC $\uparrow$} &  \multicolumn{1}{l|}{Acc}     & \multicolumn{1}{l}{FPR $\downarrow$} & \multicolumn{1}{l|}{AUC $\uparrow$}           & \multicolumn{1}{l}{FPR $\downarrow$} & \multicolumn{1}{l|}{AUC $\uparrow$}          & \multicolumn{1}{l}{FPR $\downarrow$} & \multicolumn{1}{l|}{AUC $\uparrow$}  & \multicolumn{1}{l}{FPR $\downarrow$} & \multicolumn{1}{l|}{AUC $\uparrow$}          & \multicolumn{1}{l}{FPR $\downarrow$} & \multicolumn{1}{l|}{AUC $\uparrow$} & \multicolumn{1}{l}{Acc}                        \\ \midrule
\rowcolor{lgray}  \multicolumn{15}{c}{\textit{Early fusion}} \\
DPU\textsubscript{p} & 63.7 & 78.6 & 81.2 & 70.0 & 78.4 & 71.1 & 73.9 & 56.2 & 81.8 & 38.0 & 86.2 & 59.8 & 79.8 & 72.7\\
DPU\textsubscript{f} & 61.2 & \textbf{79.3} & 81.2 & 58.8 & 81.7 & 55.9 & 82.0 & 49.6 & 83.8 & 35.9 & {87.4} & 52.3 & 82.8 & 72.7 \\
DPU\textsubscript{h} & 61.6 & 79.2 & 81.2 & 58.8 & 81.8 & 54.8 & 82.3 & 47.7 & 83.8 & 35.9 & 87.4 & 51.8 & 82.9 & 72.7 \\
\rowcolor{blue!20!white} Ours & \textbf{59.9} & 79.0 & 81.2 & \textbf{53.7} & \textbf{88.1} & \textbf{40.0} & \textbf{91.0} & \textbf{46.3} & \textbf{86.0} & \textbf{33.2} & \textbf{87.9} & \textbf{46.6} & \textbf{86.4} & 72.7\\
\midrule
\rowcolor{lgray}  \multicolumn{15}{c}{\textit{Mid fusion}} \\
DPU\textsubscript{p} & 63.9 & 78.5 & 81.0 & 69.2 & 79.6 & 70.0 & 74.5 & 53.8 & 83.3 & 33.1 & 88.8 & 58.0 & 80.9 & 74.4\\
DPU\textsubscript{f} & 62.1 & \textbf{79.0} & 81.0 & 56.3 & 83.4 & 57.5 & 81.5 & 50.9 & 84.1 & \textbf{28.3} & \textbf{90.1} & 51.0 & 83.6& 74.4\\
DPU\textsubscript{h} & 63.1 & 79.0 & 81.0 & 56.3 & 83.4 & 57.1 & 81.5 & 50.6 & 84.0 & 28.8 & \textbf{90.1} & 51.2 & 83.6 & 74.4\\
\rowcolor{blue!20!white} Ours & \textbf{60.0} & 78.7 & 81.0 & \textbf{48.3} & \textbf{89.0} & \textbf{40.2} & \textbf{90.2} & \textbf{47.6} & \textbf{85.8} & 29.9 & 89.5 & \textbf{45.2} & \textbf{86.4} & 74.4\\
\bottomrule %\bottomrule
\end{tabular}
    }
% \vspace{-0.15in}
\end{table*}

\subsection{The importance of a hybrid detector}

\begin{figure}[t]
  \centering
  \begin{minipage}{0.48\textwidth}
    \centering
    \captionof{table}{\textbf{Ablating hybrid components on Kinetics-600 Near-OOD and Kinetics-600:HAC Far-OOD.} We find that each component of our method is important. ($^\dagger$) we use the energy score~\cite{liu2020energy} for only $g(\x)$, corresponding to Eq.~\eqref{eq:logit}. 
    % ($^\ddagger$) results without scaling (Eq.~\eqref{eq:scaling}).
    }
    \label{tab:abl_comp}    
    \begin{tabular}{ccc|cc}
      \toprule
       \multirow{2}{*}{$s(\x)$} & \multirow{2}{*}{$r(\x)$} & \multirow{2}{*}{$g(\x)$} & \multicolumn{1}{c}{Near} & \multicolumn{1}{c}{Far} \\
         & & & \hspace{0.1cm} FPR \hspace{0.1cm} & FPR \hspace{0.1cm}\\
      \midrule
    \cmark & \xmark & \xmark & 64.1 & 57.1 \\
    \xmark & \cmark & \xmark & 71.6 & 52.7\\
    \xmark & \xmark & \cmark$^\dagger$ & 64.8 & 46.8\\
    \cmark & \cmark & \xmark & 68.4 & 45.9\\
    \xmark & \cmark & \cmark & 64.8 & 41.7\\
    \cmark & \xmark & \cmark & 64.6 & 47.5\\
    % \cmark$^\ddagger$ & \cmark$^\ddagger$ & \cmark & 99.9 & 50.0 & 99.9 & 50.0 \\
    \rowcolor{blue!20!white} \cmark & \cmark & \cmark & \textbf{64.0} & \textbf{41.6} \\
    \hline
    \end{tabular}
  \end{minipage}
  \hfill
  \begin{minipage}{0.48\textwidth}
    \centering
    \includegraphics[width=\linewidth]{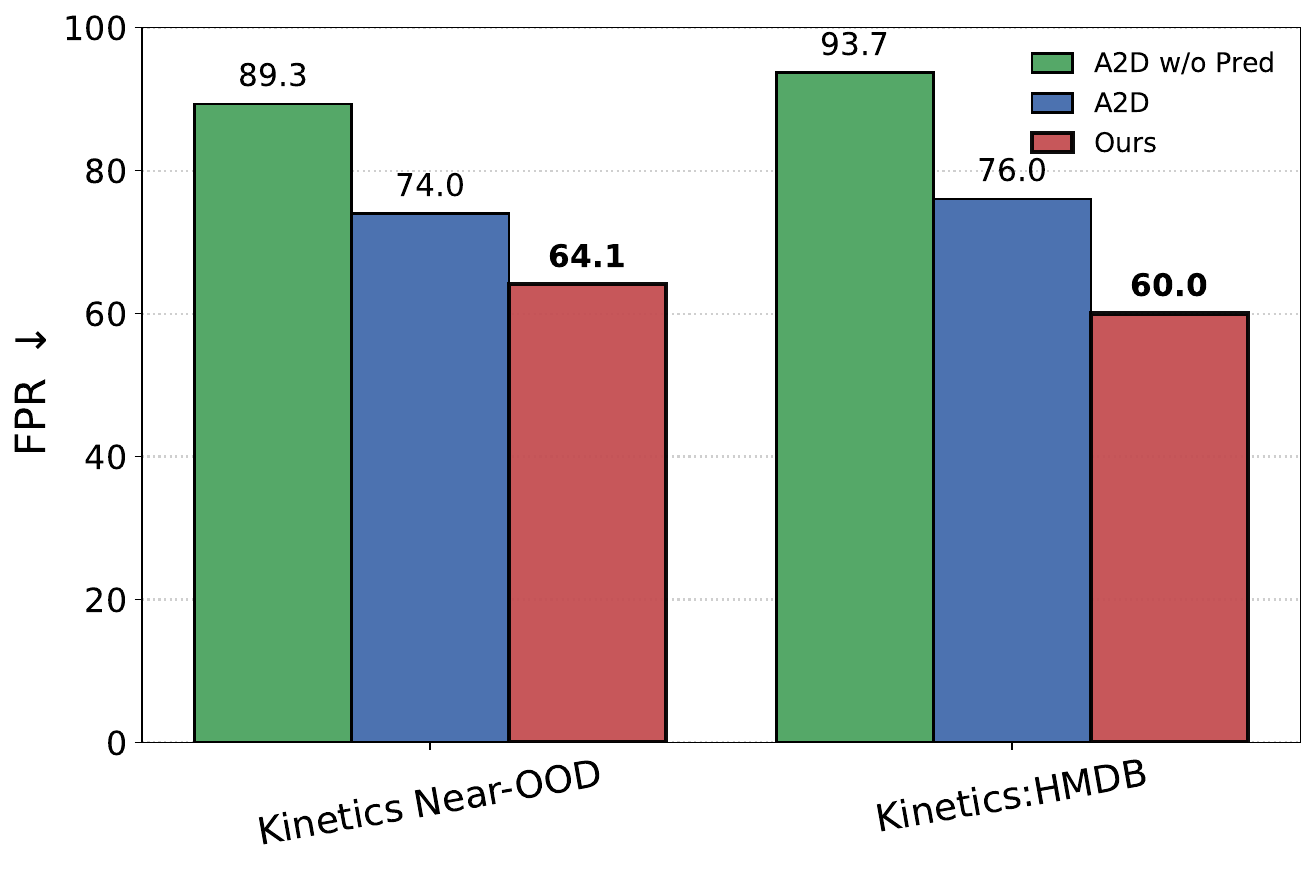}
    \captionof{figure}{\textbf{Despite being trained with A2D, the difference between uni-modal predictions does not provide a strong indicator for OOD.} Instead, we find that the relation between the predictions of the uni-modal heads and the prediction of the multi-modal head is a strong signal of normality.}
    \label{fig:a2d}
  \end{minipage}
\end{figure}

The motivation behind adopting a hybrid approach is that different sources of information provide complementary signals. As the space of OOD samples is large, unifying these sources allows the detection of a broader spectrum of OOD samples. 
To validate this, we conduct an experiment to assess the benefits of each of the three information sources we use in our detector, i.e., the multi-modal $s(\x)$, feature $r(\x)$, and probability $g(\x)$ signals, in Tab.~\ref{tab:abl_comp}. 

The results reveal several notable trends. First, while non-hybrid approaches can achieve good results in isolated cases,
they do not perform well for both types of OOD. For example, $r(\x)$ trails behind the FPR achieved by our full method on Near-OOD by $7.6$ points. Second, hybrid approaches that combine two modalities can alleviate some of these shortcomings. The combination of $s(\x)$ and $g(\x)$, for example, reaches the second-best best FPR for the Near-OOD setting. 
Nonetheless, the best results are achieved by combining all three sources, highlighting the importance of a holistic approach to OOD detection for multi-modal action recognition.

\subsection{Limitations of Modality Disagreement}
\label{sec:a2d}

The A2D method~\cite{dong2024multiood} is motivated by the observation that the distance between the predictions on individual modalities is relatively larger for OOD samples, and their regularization term aims to increase this disagreement after excluding the ground-truth class.
Despite this, A2D and related methods ignore this property at inference time, instead relying on off-the-shelf OOD detectors.
Nonetheless, A2D encodes modality disagreement as a signal for OOD in two ways, which we refer to as \textit{A2D}, the distance between the full uni-modal probability distributions, and \textit{A2D w/o Pred}, the distance between uni-modal predictions excluding the predicted class.
To test their efficacy, we evaluate both variants as OOD detectors and compare them to our proposed score $s(\x)$. 
As shown in Fig.~\ref{fig:a2d}, the discrepancy between uni-modal predictions alone is a weak indicator of OOD. In contrast, our score based on the relation between uni- and multi-modal predictions provides a stronger signal, improving the FPR on Kinetics Near-OOD by $9.9$ FPR and on Kinetics:HMDB Far-OOD by $16.0$ FPR.
% These findings underscore the need for a more principled modality-aware scoring function, which we address. 

\begin{figure*}[t]
  \centering
  \setlength\tabcolsep{1pt}
  \begin{tabular}{ccc}
    \includegraphics[width=0.33\linewidth]{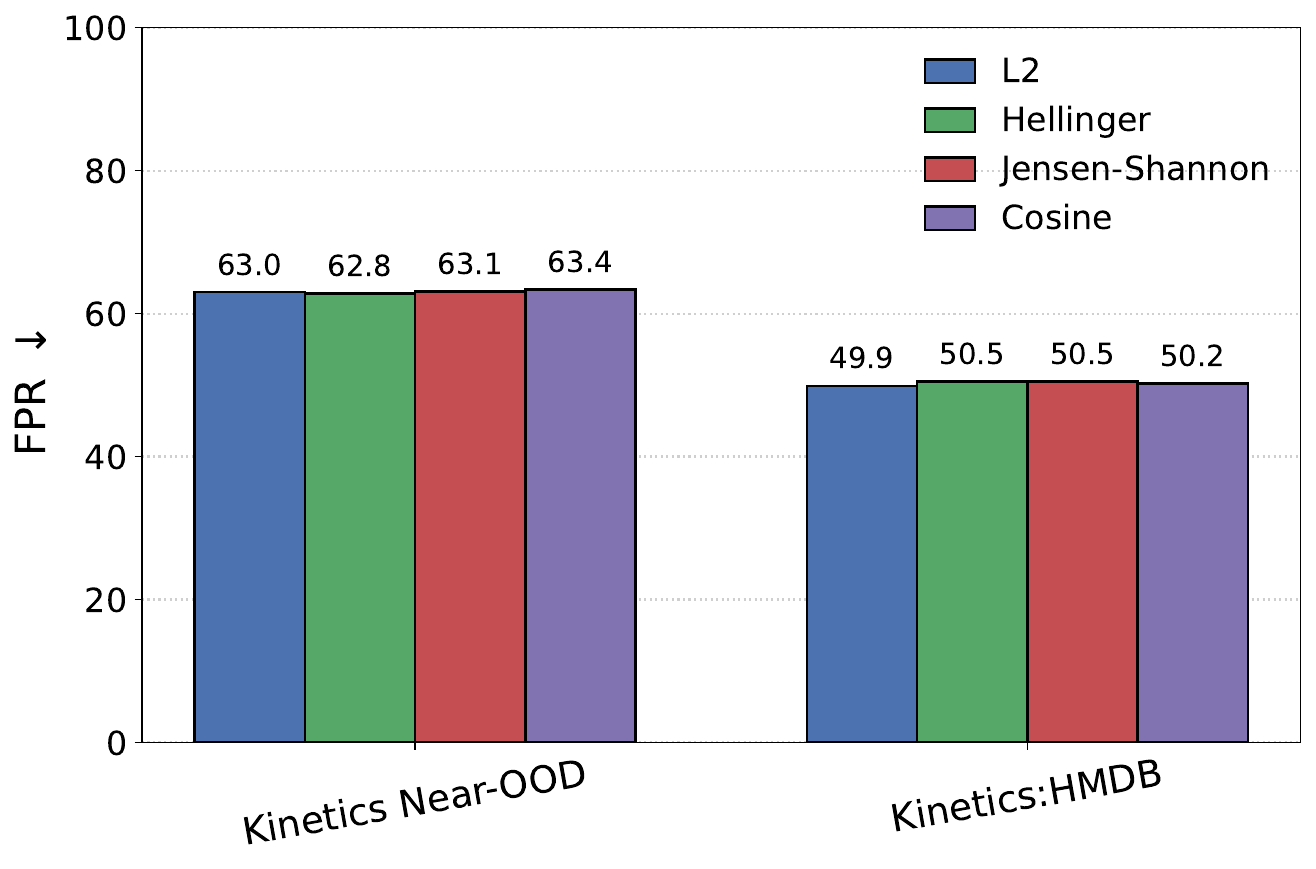} &
    \includegraphics[width=0.33\linewidth]{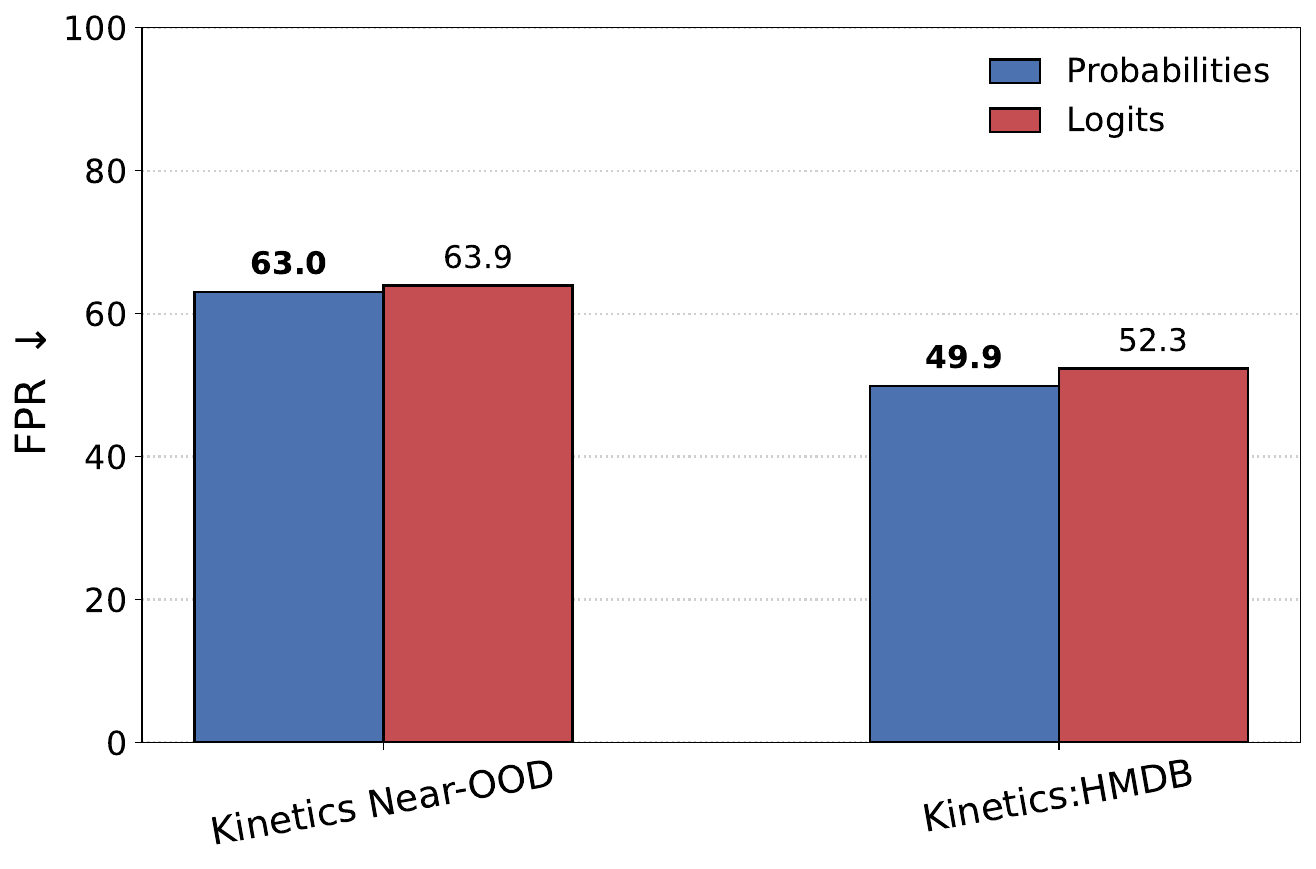} &
    \includegraphics[width=0.33\linewidth]{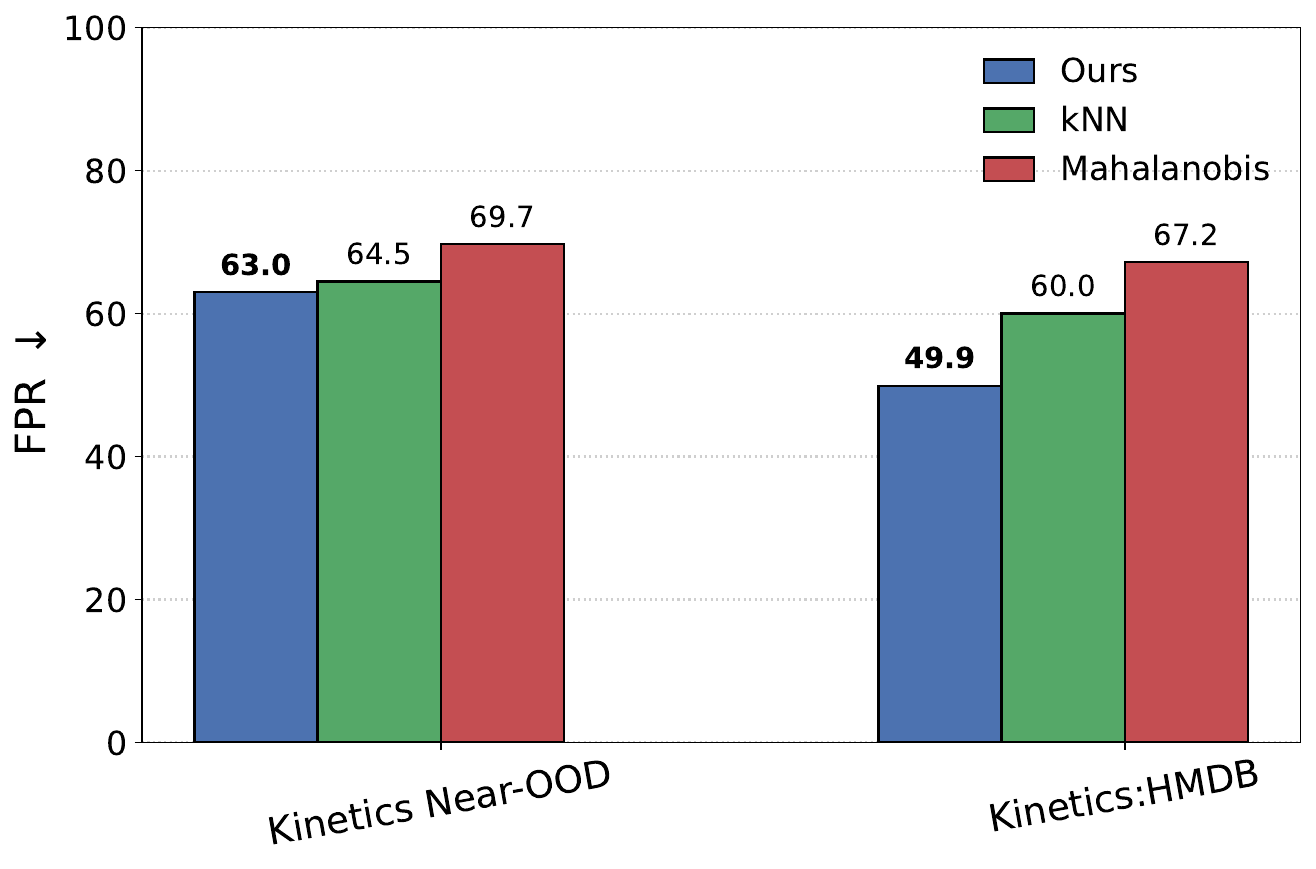} \\
    (a) & (b) & (c) \\
  \end{tabular}
    \caption{\textbf{Ablation studies for our framework.} We evaluate the impact of (a) the distance function, (b) the choice for probabilities over logits in Eq.~\eqref{eq:dist}, and (c) variants of feature-based scores for $r(\x)$. 
    }
    \label{fig:ablations}
\end{figure*}

\subsection{Ablations}

\textbf{Complexity of $w$.}
We ablate the choice for our linear system (``class-conditional") in Eq.~\eqref{eq:linear} with respect to other possible versions. Specifically, we compare to a fully connected $\mathbf{W} \in \mathbb{R}^{(MC)\times C}$ (``fully connected"), where every uni-modal logit influences every multi-modal logit, and a version with only a single $\tilde{\mathbf{w}}$ shared between all classes (``single linear").
We show the results in Tab.~\ref{tab:abl}(left), where we find that the overall performance is stable, with the class-conditional model slightly outperforming the other two variants.

\begin{figure}[t]
  \centering
  \setlength\tabcolsep{1pt}
  \begin{tabular}{ccc}
    \includegraphics[width=0.32\linewidth]{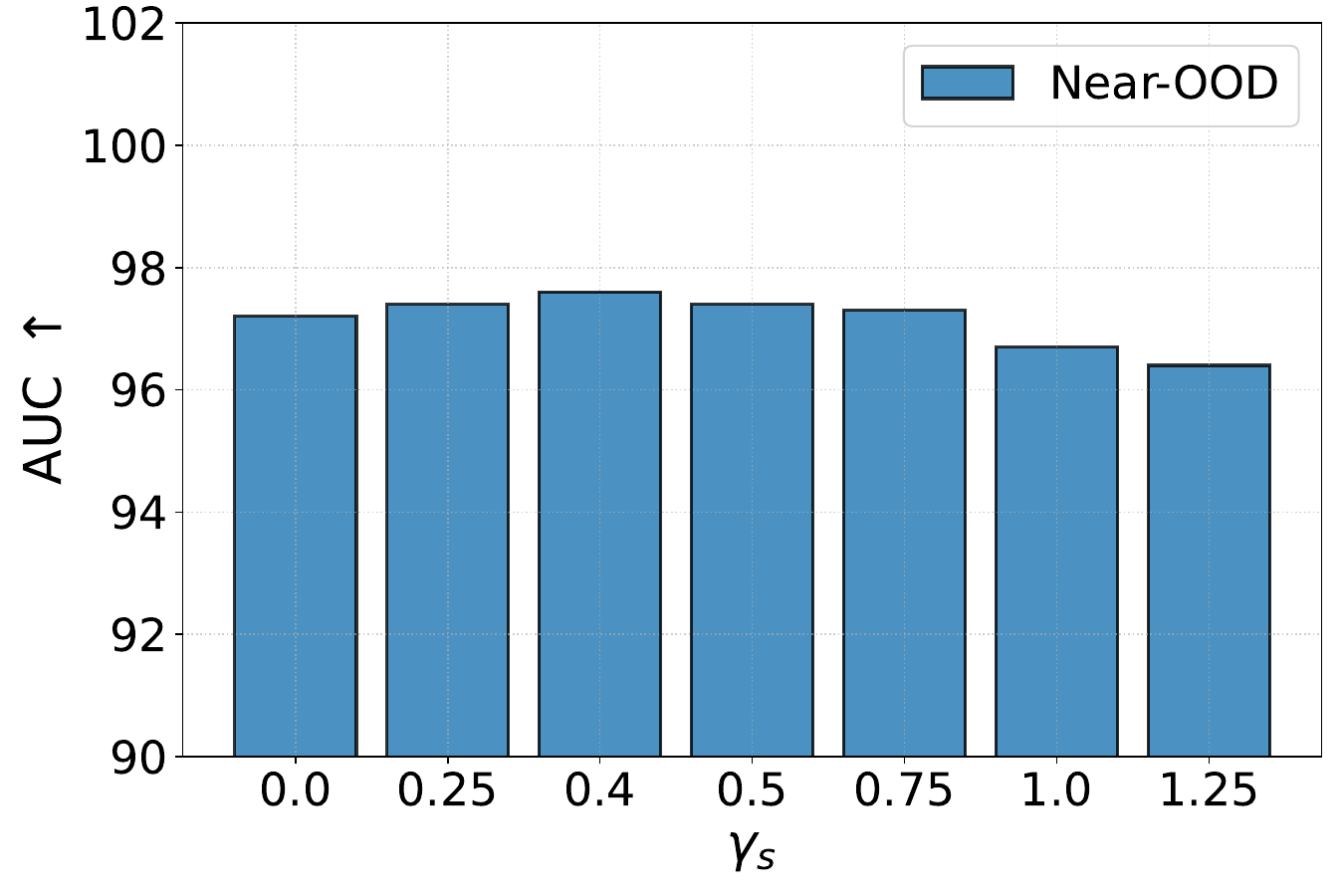} &
    \includegraphics[width=0.32\linewidth]{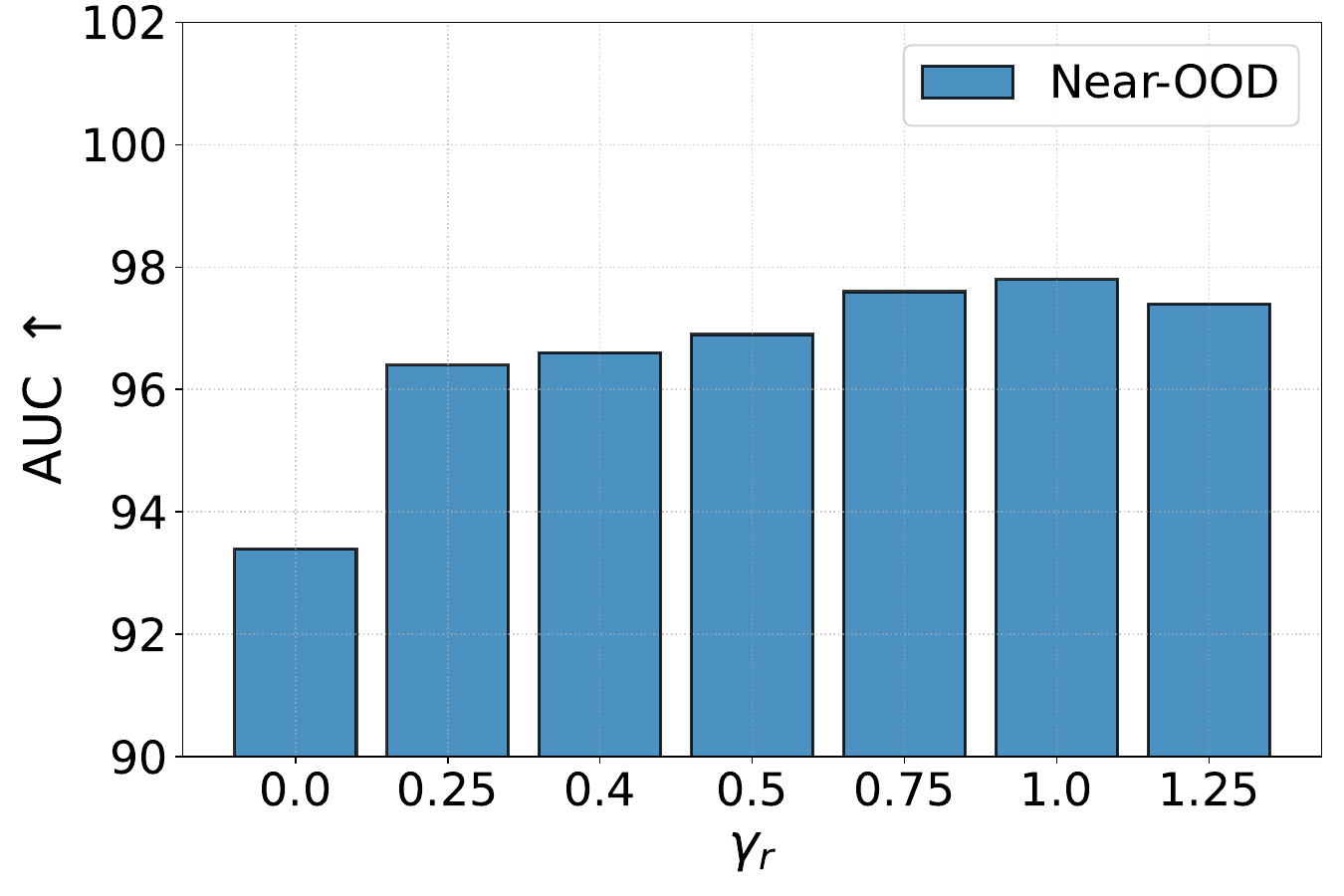} &
    \includegraphics[width=0.32\linewidth]{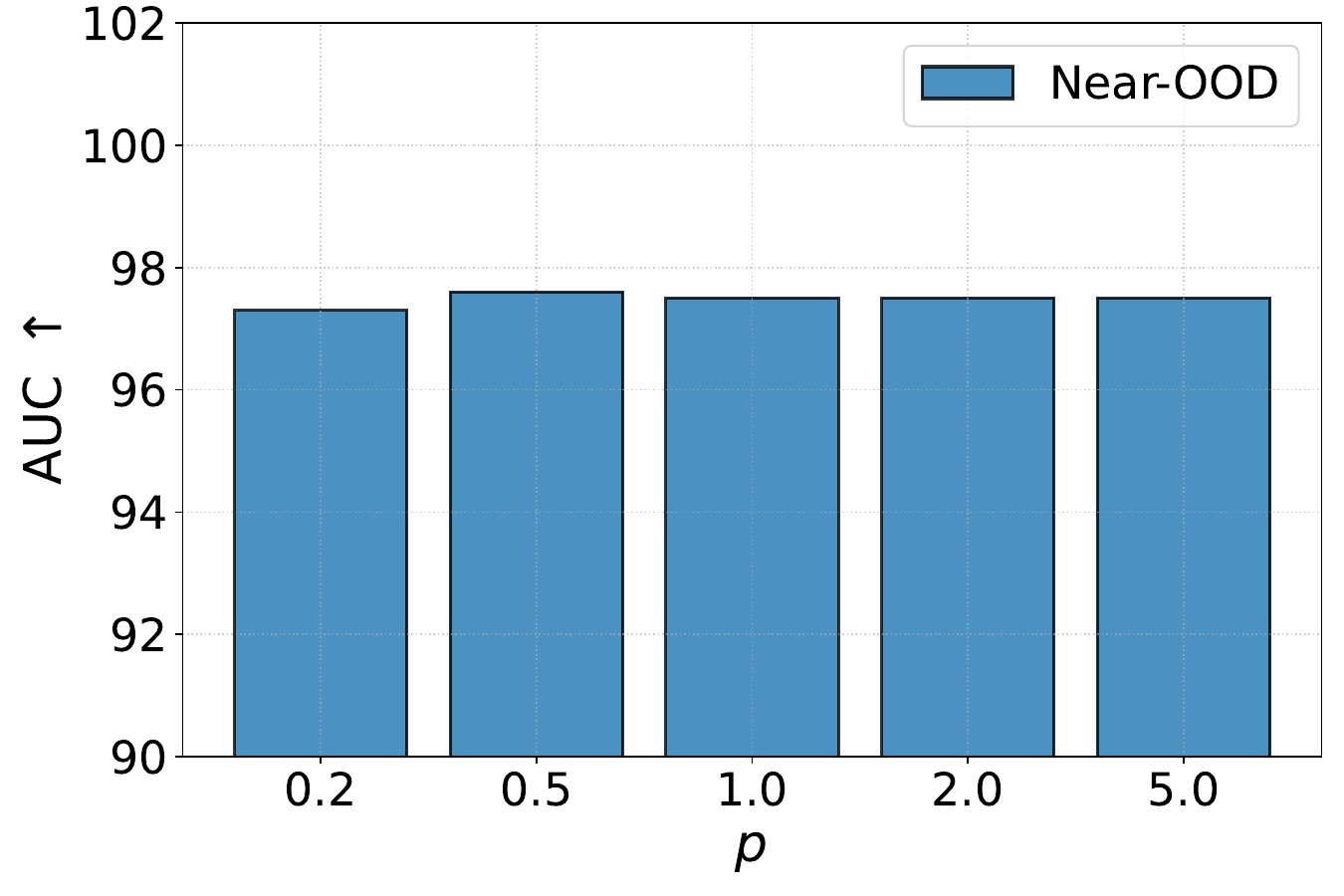} \\
    (a) & (b) & (c) \\
  \end{tabular}
    \caption{\textbf{Sensitivity to hyperparameters.} We show the AUC for HMDB51:Kinetics Far-OOD for different settings of the hyperparameters a) $\gamma_s$, b) $\gamma_r$, and c) $p$. Our method is robust to their settings.
    }
    \label{fig:sensitivity}
\end{figure}

\begin{table}[t]
\begin{center}
\caption{\textbf{Ablating $w$ complexity and validation set size.} \textit{Left:} We report the results for Kinetics Near-OOD with early and late fusion architectures. Our class-conditional approach gives the best balance. \textit{Right:} Robustness of results with respect to subsampling the validation set for Kinetics Near-OOD and Kinetics:HMDB. Our results remain stable even when only using a fraction of the validation samples.}
 \begin{tabular}{c|cccc}
    \toprule
      & \multicolumn{2}{c}{Early} & \multicolumn{2}{c}{Late} \\
       & FPR & AUC & FPR & AUC \\
    \midrule
    % Single weight & 59.5 & 87.5 & 0.534\\
    Single linear & 59.9 & 79.0 & 63.3 & 77.1 \\   
    \rowcolor{blue!20!white} Class-conditional  & 59.9 & 79.0 & 63.0 & 77.0\\
    Fully connected  & 59.9 & 79.0 & 63.4 & 77.0 \\
   \hline
\end{tabular}
\quad\quad
     \begin{tabular}{c|cccc}
        \toprule
         & \multicolumn{2}{c}{Near} & \multicolumn{2}{c}{Far} \\
       & FPR & AUC & FPR & AUC \\
        \hline
         0.5\% & 63.3 & 76.5  & 50.5& 88.0\\
         1\% & 63.2 & 77.2  & 50.8 & 88.3\\
         5\% & 63.0 & 77.0 & 50.5 & 88.4\\
         10\% & 63.0 & 77.1 & 50.3 & 88.5 \\
         Full & 63.0 & 77.0 & 49.9 & 88.5\\
        \hline
    \end{tabular}%
\label{tab:abl}
% }
\end{center}
\end{table}

\textbf{Other choices.}
We ablate the choice of distance function (Eq.~\eqref{eq:dist}) in Fig.~\ref{fig:ablations}(a). The differences between the distance functions are marginal: out of the four distances considered, the maximum difference in FPR on the two experiments is $0.6$.
In Fig.~\ref{fig:ablations}(b), we justify our choice to use the distance between probability distributions rather than logits. In general, the logits for ID samples have a higher norm, which makes the distance-based comparison imbalanced. By considering probability distributions instead, the resulting values for ID and OOD samples have the same range, leading to better performance.
Finally, we compare our feature-based score $r(\x)$ with the kNN distance~\cite{sun2022out} and the class-wise Mahalanobis method~\cite{lee2018mahalanobis}.
The results in Fig.~\ref{fig:ablations}(c) confirm that the choice for invariants results in the best performance.

\textbf{Sensitivity to validation set size.}
We test the sensitivity of our method to the validation set size and show the results in Tab.~\ref{tab:abl}(right). We do not find a strong degradation in performance; even when only using $1$\% of the validation set, the performance of our overall model remains within $0.2$ points of the AUC obtained with the full validation set. This is a direct consequence of only estimating the intercept plus one parameter per modality for each class.

\textbf{Sensitivity to hyperparameters.}
Finally, we probe the effect of the hyperparameters of our method: $\gamma_s$, $\gamma_r$, and $p$.
We present the results in Fig.~\ref{fig:sensitivity},
which demonstrates that our method is generally robust to their settings.

\textbf{Limitations.}
Our method inherits a limitation of the broader multi-modal OOD field: a reliance on architectures with exposed uni-modal classifiers. However, we go beyond previous works in this domain by showing that our method also works under various fusion mechanisms. Moreover, we show in the supplementary material that our method still performs well when adding uni-modal heads post hoc to an already trained classifier.
We show qualitative examples of failure cases in the supplementary material.

\section{Conclusion}

We presented the first modality-aware OOD detector for multi-modal action recognition. Our approach is based on the observation that the probability distribution predicted by the multi-modal classifier can be modeled as an aggregation of the uni-modal predictions. The residual error between the observed and predicted distributions is higher for OOD samples, leading to a novel modality-aware OOD scoring function.
We integrate this score with information from the feature space by modeling the invariants in the multi-modal manifold and the final predictions as a virtual logit. 
Through combining these complementary signals, we outperform the average state of the art on the challenging MultiOOD benchmark.
Overall, our findings demonstrate that our approach can enhance the robustness of multi-modal action recognition systems. As our approach does not make any modality-specific assumptions, we expect it to generalize well to other domains. We provide an initial exploration of this in the supplementary material.

\section*{Acknowledgements}
The work has been supported by the ERC Consolidator Grant FORHUE (101044724).
For the computations involved in this research, we acknowledge EuroHPC Joint Undertaking for awarding us access to Leonardo at CINECA, Italy, through EuroHPC Regular Access Call - proposal No. EHPC-REG-2025R01-218.

%Please insert your acknowledgments here.

\bibliographystyle{splncs04}
\bibliography{refs}

\clearpage
\setcounter{page}{1}
% \maketitlesupplementary

\section{Full Experimental Details}

We provide more details on the dataset combinations used for the experiments. All datasets follow the setup and pre-processing of the Multi-OOD benchmark~\cite{dong2024multiood}.
\begin{itemize}
    \item \textbf{Kinetics-600 Far-OOD.} The in-distribution consists of 229 classes that do not overlap with the OOD datasets used. The OOD datasets considered are UCF101, EPIC-Kitchen, HAC, and HMDB51.
    \item \textbf{HMDB51 Far-OOD.} The in-distribution consists of 43 classes that do not overlap with the OOD datasets used. The OOD datasets considered are UCF101, EPIC-Kitchen, HAC, and Kinetics-600.
    \item \textbf{Near-OOD.} We consider four near-OOD experiments: EPIC-Kitchen, split into 4 classes as ID and 4 as OOD, HMDB51, split into 25 classes as ID 26 as OOD, UCF101, split into 50 classes as ID and 51 as OOD, and Kinetics-600, split into 129 classes as ID and 100 as OOD.
    \item \textbf{Triple modal.} For Near-OOD, we use the 4 classes of EPIC-Kitchens as ID, with the other 4 as OOD. For Far-OOD, we use the same data as ID, and use Kinetics as OOD. Note that some of the datasets, e.g., HMDB and UCF, only provide two modalities and can therefore not be used in this setting.
\end{itemize}

\subsection{ID accuracies for Near-OOD experiments}

We show the ID accuracies of the classifiers used in Tab.~\ref{tab:near_accs}. Our OOD detector is applied to the exact same network as the DPU variants, showing its effectiveness over single-modal detectors. Furthermore, the accuracies obtained by the A2D classifiers are higher on both HMDB and Kinetics. This demonstrates that our gains are not just due to using a better-performing classifier, but actually stem from the OOD detector itself.

\begin{table*}[b]
%vspace{0.1in}
\caption{\textbf{ID accuracies for the multimodal Near-OOD Detection results using video and optical flow.} The ID accuracies are similar between methods and are not responsible for differences in OOD detection performance. }
\label{tab:near_accs}
% \vspace{-0.1in}
\centering
% \renewcommand{\arraystretch}{1.00}
% \setlength{\tabcolsep}{7pt} 
% \fontsize{8}{10}\selectfont 
% \resizebox{0.85\textwidth}{!}{
\begin{tabular}{r|cccc}
\toprule %\toprule
\multicolumn{1}{c|}{\multirow{2}{*}{Method}} & \multicolumn{4}{c|}{Datasets}                                        \\ \cline{2-5}
\multicolumn{1}{c|}{}                        & HMDB51                                 & \multicolumn{1}{c|}{UCF101}                                     & \multicolumn{1}{c|}{Kinetics-600}                                        & \multicolumn{1}{c|}{EPIC-Kitchen}          \\ \cline{2-5}
% \multicolumn{1}{c|}{}                        & \multicolumn{1}{l}{FPR $\downarrow$} & \multicolumn{1}{l|}{AUC $\uparrow$}           & \multicolumn{1}{l}{FPR $\downarrow$} & \multicolumn{1}{l|}{AUC $\uparrow$}          & \multicolumn{1}{l}{FPR $\downarrow$} & \multicolumn{1}{l|}{AUC $\uparrow$}     & \multicolumn{1}{l}{FPR $\downarrow$} & \multicolumn{1}{l|}{AUC $\uparrow$}       & \multicolumn{1}{l}{FPR $\downarrow$} & \multicolumn{1}{l|}{AUC $\uparrow$}                   \\ 
\midrule
A2D\textsubscript{p} & 91.1 & 99.7 & 81.6 & 72.5\\
A2D\textsubscript{f} & 91.1 & 99.7 & 81.6 & 72.5\\
A2D\textsubscript{h} & 91.1 & 99.7 & 81.6 & 72.5\\
DPU\textsubscript{p} & 90.8 & 99.7 & 81.6 & 71.6 \\
DPU\textsubscript{f} & 90.8 & 99.7 & 81.6 & 71.6 \\
DPU\textsubscript{h} & 90.8 & 99.7 & 81.6 & 71.6 \\
\rowcolor{blue!20!white} Ours & 90.8 & 99.7 & 81.6 & 71.6 \\
\bottomrule %\bottomrule
\end{tabular}
    % }
% \vspace{-0.15in}
\end{table*}

\subsection{Detectors used in main results}

We evaluate our method against the strongest reported versions of previous multi-modal OOD methods: A2D with NP-Mix (A2D)~\cite{dong2024multiood}, DPU~\cite{li2025dpu}, and FM~\cite{liu2025extremely}. As the A2D and DPU methods are evaluated using a variety of off-the-shelf detectors, we compare against the best-performing variant of each baseline method across three detector categories: probabilities/logits-based, feature-based, and hybrid versions. We denote these throughout the paper with subscripts \textsubscript{p}, \textsubscript{f}, and \textsubscript{h}, respectively. Thus, A2D\textsubscript{f} denotes the best results of A2D with a feature-based scoring function. Here, we give the exact method used for every baseline in Tabs.~\ref{tab:kin_far_methods}-\ref{tab:near_methods}. We find that, for the baselines, different detectors perform best in different scenarios. Notably, our proposed method always uses the same settings and achieves the best results.

\begin{table*}[t]
\caption{\textbf{Scoring functions used per baseline in the main results for multimodal Far-OOD Detection with Kinetics-600 as the ID dataset.} We selected the best-performing method per category for each experiment.}
\label{tab:kin_far_methods}
% \vspace{-0.1in}
\centering
% \renewcommand{\arraystretch}{1.00}
% \setlength{\tabcolsep}{7pt} 
% \fontsize{8}{10}\selectfont 
% \resizebox{0.9\textwidth}{!}{
\begin{tabular}{r|cccc}
\toprule %\toprule
\multicolumn{1}{c|}{\multirow{2}{*}{Method}} & \multicolumn{4}{c|}{OOD Datasets}  \\ \cline{2-5}
\multicolumn{1}{c|}{}                        & \multicolumn{1}{c|}{HMDB51}                                 & \multicolumn{1}{c|}{UCF101}                                     & \multicolumn{1}{c|}{HAC}                                        & \multicolumn{1}{c|}{EPIC-Kitchen}                     \\  \midrule
A2D\textsubscript{p} & GEN & GEN & GEN & GEN \\
A2D\textsubscript{f} & ASH & ASH & ASH & ASH \\
A2D\textsubscript{h} & ViM & ViM & ViM & ViM \\
DPU\textsubscript{p} & GEN & GEN & GEN & GEN \\
DPU\textsubscript{f} & KNN & KNN & KNN & KNN \\
DPU\textsubscript{h} & ViM & ViM & ViM & ViM \\
\bottomrule %\bottomrule
\end{tabular}
    % }
% \vspace{-0.15in}
\end{table*}

\begin{table*}[t]
\caption{\textbf{Scoring functions used per baseline in the main results for multimodal Far-OOD Detection with HMDB as the ID dataset.} We selected the best-performing method per category for each experiment. }
\label{tab:hmdb_far_methods}
% \vspace{-0.1in}
\centering
% \renewcommand{\arraystretch}{1.00}
% \setlength{\tabcolsep}{7pt} 
% \fontsize{8}{10}\selectfont 
% \resizebox{0.9\textwidth}{!}{
\begin{tabular}{r|cccc}
\toprule %\toprule
\multicolumn{1}{c|}{\multirow{2}{*}{Method}} & \multicolumn{4}{c|}{OOD Datasets}  \\ \cline{2-5}
\multicolumn{1}{c|}{}                        & \multicolumn{1}{c|}{Kinetics-600}                                 & \multicolumn{1}{c|}{UCF101}                                     & \multicolumn{1}{c|}{HAC}                                        & \multicolumn{1}{c|}{EPIC-Kitchen}                     \\  \midrule
A2D\textsubscript{p} & Energy & Energy & Energy & Energy \\
A2D\textsubscript{f} & Mahalanobis & Mahalanobis & Mahalanobis & Mahalanobis \\
A2D\textsubscript{h} & ViM & ViM & ViM & ViM \\
DPU\textsubscript{p} & GEN & GEN & GEN & GEN \\
DPU\textsubscript{f} & Mahalanobis & Mahalanobis & Mahalanobis & Mahalanobis \\
DPU\textsubscript{h} & ViM & ViM & ViM & ViM \\
\bottomrule %\bottomrule
\end{tabular}
    % }
% \vspace{-0.15in}
\end{table*}

\begin{table*}[t]
\caption{\textbf{Scoring functions used per baseline in the main results for multimodal Near-OOD Detection.} We selected the best-performing method per category for each experiment.}
\label{tab:near_methods}
% \vspace{-0.1in}
\centering
% \renewcommand{\arraystretch}{1.00}
% \setlength{\tabcolsep}{7pt} 
% \fontsize{8}{10}\selectfont 
% \resizebox{0.9\textwidth}{!}{
\begin{tabular}{r|cccc}
\toprule %\toprule
\multicolumn{1}{c|}{\multirow{2}{*}{Method}} & \multicolumn{4}{c|}{OOD Datasets}  \\ \cline{2-5}
\multicolumn{1}{c|}{}                        & \multicolumn{1}{c|}{HMDB51}                                 & \multicolumn{1}{c|}{UCF101}                                     & \multicolumn{1}{c|}{Kinetics-600}                                        & \multicolumn{1}{c|}{EPIC-Kitchen}                     \\  \midrule
A2D\textsubscript{p} & GEN & MSP & GEN & GEN \\
A2D\textsubscript{f} & KNN & KNN & KNN & KNN \\
A2D\textsubscript{h} & ViM & ViM & ViM & ViM \\
DPU\textsubscript{p} & MSP & MSP & MSP & MSP \\
DPU\textsubscript{f} & KNN & KNN & KNN & KNN \\
DPU\textsubscript{h} & ViM & ViM & ViM & ViM \\
\bottomrule %\bottomrule
\end{tabular}
    % }
% \vspace{-0.15in}
\end{table*}

\subsection{Linear correlation for all experiments}
Tab.~\ref{tab:corr} shows that the linear correlation holds across all experiments considered. In general, it is highest for late-fusion architectures, but also holds for non-linear early- and mid-fusion mechanisms. 

\begin{table}[t]
    \centering
    % \resizebox{\linewidth}{!}{
    \caption{\textbf{The linearity assumption holds across datasets and non-linear fusion using cross-attention (CA).} We show the Pearson correlation, which remains consistently high.
    }
    \label{tab:corr}%
    \begin{tabular}{cccccc}
        \toprule
         & non-lin. & HMDB & UCF & Kinetics & EPIC\\
        \hline
         Early (CA) & $\checkmark$ & 94.9 & 91.7 & 92.6 & 98.4 \\
         Mid (CA) & $\checkmark$ & 97.1 & 91.6 & 93.4 & 98.1 \\
         Late     &          & 99.8 & 99.9 & 99.8 & 99.8\\
        \hline
    \end{tabular}%
\end{table}

\subsection{Computational cost}
We compare inference time on the full HMDB:Kinetics experiment to the strongest baselines in Tab.~\ref{tab:cost}, using one RTX 3090.  
The inference overhead of the OOD detectors remains low.

\begin{table}[t]
    \centering
    % \resizebox{0.9\linewidth}{!}{
    \caption{\textbf{Inference time comparison.} We report the mean wall clock time over five runs for the full HMDB:Kinetics experiment. Computational cost for our method is similar to the other baselines.}
    % }
    \label{tab:cost}%
    \begin{tabular}{c|ccccc}
        \toprule
         & A2D/DPU\textsubscript{GEN} & A2D/DPU\textsubscript{Maha} & A2D/DPU\textsubscript{ViM} & Ours \\
        \hline
        % Time (s) & 0.013 & 27.042 & 14.796 & 9.726\\
        Time (s) & 920.0 & 947.0 & 934.8 & 929.7 \\
        \hline
    \end{tabular}%
\end{table}

\subsection{Comparison to published results}

Previous works only provided results for a single run. To assess the variability in results, we re-trained the baselines using their official code and settings over three random seeds. We compare our results to those reported in the respective publications to validate our setup. 
We show the differences between reported and our results for A2D\textsubscript{p} and DPU\textsubscript{p} in Tab.~\ref{tab:comp_kin_far}, Tab.~\ref{tab:comp_hmdb_far}, and Tab.~\ref{tab:comp_near}. Overall, the numbers we obtain are very similar to those reported in their official publications.

\begin{table*}[]
\caption{\textbf{Similarity between reported and reproduced results for Far-OOD with Kinetics as ID.} }
\label{tab:comp_kin_far}
% \vspace{-0.1in}
\centering
% \renewcommand{\arraystretch}{1.00}
% \setlength{\tabcolsep}{7pt} 
% \fontsize{8}{10}\selectfont 
\resizebox{0.85\textwidth}{!}{
\begin{tabular}{r|cccccccc|ccc}
\toprule %\toprule
\multicolumn{1}{c|}{\multirow{3}{*}{Method}} & \multicolumn{10}{c|}{OOD Datasets}                                       & \multicolumn{1}{l}{\multirow{3}{*}{Acc}} \\ \cline{2-11}
\multicolumn{1}{c|}{}                        & \multicolumn{2}{c|}{HMDB51}                                 & \multicolumn{2}{c|}{UCF101}                                     & \multicolumn{2}{c|}{HAC}                                        & \multicolumn{2}{c|}{EPIC-Kitchen}  & \multicolumn{2}{c|}{Average}                      & \multicolumn{1}{l}{}                        \\ \cline{2-11}
\multicolumn{1}{c|}{}                        & \multicolumn{1}{l}{FPR $\downarrow$} & \multicolumn{1}{l|}{AUC $\uparrow$}           & \multicolumn{1}{l}{FPR $\downarrow$} & \multicolumn{1}{l|}{AUC $\uparrow$}          & \multicolumn{1}{l}{FPR $\downarrow$} & \multicolumn{1}{l|}{AUC $\uparrow$}  & \multicolumn{1}{l}{FPR $\downarrow$} & \multicolumn{1}{l|}{AUC $\uparrow$}          & \multicolumn{1}{l}{FPR $\downarrow$} & \multicolumn{1}{l|}{AUC $\uparrow$} & \multicolumn{1}{l}{}                        \\ \midrule
DPU\textsubscript{p} & 53.9 & 84.5 & 45.6 & 85.6 & 49.1 & 83.3 & 39.9 & 85.7 & 47.1 & 84.8 & 76.7\\
DPU\textsubscript{p} & 54.0\tiny{$\pm$2.4} & 83.2\tiny{$\pm$0.4} & 52.6\tiny{$\pm$1.5} & 81.8\tiny{$\pm$0.4} & 46.6\tiny{$\pm$2.4} & 84.8\tiny{$\pm$1.2} & 30.5\tiny{$\pm$0.5} & 89.7\tiny{$\pm$0.9} & 45.9 & 84.9 & 75.2\\
\bottomrule %\bottomrule
\end{tabular}
    }
% \vspace{-0.15in}
\end{table*}

\begin{table*}
\caption{\textbf{Similarity between reported and reproduced results for Far-OOD with HMDB as ID.} }
\label{tab:comp_hmdb_far}
% \vspace{-0.1in}
\centering
% \renewcommand{\arraystretch}{1.00}
% \setlength{\tabcolsep}{7pt} 
% \fontsize{8}{10}\selectfont 
\resizebox{0.85\textwidth}{!}{
\begin{tabular}{r|cccccccccc|c}
\toprule %\toprule
\multicolumn{1}{c|}{\multirow{3}{*}{Method}} & \multicolumn{10}{c|}{OOD Datasets}                                       & \multicolumn{1}{l}{\multirow{3}{*}{Acc}} \\ \cline{2-11}
\multicolumn{1}{c|}{}                        & \multicolumn{2}{c|}{Kinetics-600}                                 & \multicolumn{2}{c|}{UCF101}                                     & \multicolumn{2}{c|}{HAC}                                        & \multicolumn{2}{c|}{EPIC-Kitchen}         & \multicolumn{2}{c|}{Average}               & \multicolumn{1}{l}{}                        \\ \cline{2-11}
\multicolumn{1}{c|}{}                        & \multicolumn{1}{l}{FPR $\downarrow$} & \multicolumn{1}{l|}{AUC $\uparrow$}           & \multicolumn{1}{l}{FPR $\downarrow$} & \multicolumn{1}{l|}{AUC $\uparrow$}          & \multicolumn{1}{l}{FPR $\downarrow$} & \multicolumn{1}{l|}{AUC $\uparrow$}          & \multicolumn{1}{l}{FPR $\downarrow$} & \multicolumn{1}{l|}{AUC $\uparrow$}  & \multicolumn{1}{l}{FPR $\downarrow$} & \multicolumn{1}{l|}{AUC $\uparrow$} & \multicolumn{1}{l}{}                        \\ 
\midrule
A2D\textsubscript{p} & 24.5 & 94.0 & 36.5 & 89.7 & 23.0 & 94.4 & 7.0 & 97.5 & 22.7 & 93.9 & 86.9 \\
A2D\textsubscript{p} & 24.1\tiny{$\pm$5.7} & 94.3\tiny{$\pm$1.3} & 35.5\tiny{$\pm$2.7} & 90.2\tiny{$\pm$0.1} & 22.7\tiny{$\pm$4.4} & 94.6\tiny{$\pm$0.9} & 9.5\tiny{$\pm$5.1} & 97.4\tiny{$\pm$1.1} & 23.0 & 94.1 & 86.9 \\
DPU\textsubscript{p} & 7.2 & 98.4 & 36.4 & 90.9 & 21.4 & 95.7 & 3.2 & 99.3 & 17.1 & 96.1 & 87.3 \\
DPU\textsubscript{p} & 19.3\tiny{$\pm$0.2} & 96.0\tiny{$\pm$0.4} & 27.2\tiny{$\pm$1.9} & 93.7\tiny{$\pm$0.5} & 14.8\tiny{$\pm$1.3} & 96.9\tiny{$\pm$0.4} & 6.3\tiny{$\pm$7.0} & 98.7\tiny{$\pm$1.2} & 16.9 & 96.3 & 87.4\\
\bottomrule %\bottomrule
\end{tabular}
    }
% \vspace{-0.15in}
\end{table*}

\begin{table*}
\caption{\textbf{Similarity between reported and reproduced results for Near-OOD.} }
\label{tab:comp_near}
% \vspace{-0.1in}
\centering
% \renewcommand{\arraystretch}{1.00}
% \setlength{\tabcolsep}{7pt} 
% \fontsize{8}{10}\selectfont 
\resizebox{0.85\textwidth}{!}{
\begin{tabular}{r|cccccccccc}
\toprule %\toprule
\multicolumn{1}{c|}{\multirow{2}{*}{Method}} & \multicolumn{10}{c|}{Datasets}                                        \\ \cline{2-11}
\multicolumn{1}{c|}{}                        & \multicolumn{2}{c|}{HMDB51}                                 & \multicolumn{2}{c|}{UCF101}                                     & \multicolumn{2}{c|}{Kinetics-600}                                        & \multicolumn{2}{c|}{EPIC-Kitchen}          & \multicolumn{2}{c|}{Average}                                 \\ \cline{2-11}
\multicolumn{1}{c|}{}                        & \multicolumn{1}{l}{FPR $\downarrow$} & \multicolumn{1}{l|}{AUC $\uparrow$}           & \multicolumn{1}{l}{FPR $\downarrow$} & \multicolumn{1}{l|}{AUC $\uparrow$}          & \multicolumn{1}{l}{FPR $\downarrow$} & \multicolumn{1}{l|}{AUC $\uparrow$}     & \multicolumn{1}{l}{FPR $\downarrow$} & \multicolumn{1}{l|}{AUC $\uparrow$}       & \multicolumn{1}{l}{FPR $\downarrow$} & \multicolumn{1}{l|}{AUC $\uparrow$}                   \\ \midrule
A2D\textsubscript{p} & 36.0 & 89.8 & 7.7 & 98.3 & 63.0 & 77.0 & 65.3 & 75.2 & 43.0 & 85.1 \\
A2D\textsubscript{p} & 38.4\tiny{$\pm$3.4} & 89.1\tiny{$\pm$0.7} & 8.0\tiny{$\pm$1.8} & 98.3\tiny{$\pm$0.2} & 61.2\tiny{$\pm$2.9} & 77.4\tiny{$\pm$0.7} & 70.0\tiny{$\pm$4.9} & 71.7\tiny{$\pm$3.4} & 44.4 & 84.1 \\
DPU\textsubscript{p} & 34.2 & 89.2 & 7.6 & 98.2 & 61.6 & 77.5 & 63.8 & 71.5 & 41.8 & 84.1 \\
DPU\textsubscript{p} & 35.9\tiny{$\pm$2.4} & 89.0\tiny{$\pm$0.4} & 10.0\tiny{$\pm$4.6} & 97.7\tiny{$\pm$0.8} & 63.4\tiny{$\pm$0.2} & 76.9\tiny{$\pm$0.7} & 70.7\tiny{$\pm$2.1} & 68.5\tiny{$\pm$0.7} & 45.0 & 83.0 \\
\bottomrule %\bottomrule
\end{tabular}
    }
% \vspace{-0.15in}
\end{table*}

\section{Post-hoc addition of uni-modal heads}
Our method relies on exposed uni-modal heads to derive its modality-aware score. 
However, not all multi-modal pipelines include these.
Here, we add uni-modal heads to already-trained classifiers to demonstrate the broader applicability of our approach. Specifically, we trained a network on HMDB and UCF without uni-modal heads for 50 epochs, then added uni-modal heads and fine-tuned it for one epoch. The results in Tab.~\ref{tab:posthoc} show that our method still reaches strong performance when the uni-modal heads are added post hoc, demonstrating that uni-modal heads are not a limitation in practice. 

\begin{table}[t]
    \centering
    % \resizebox{\textwidth}{!}{
    \caption{\textbf{Success with post-hoc addition of uni-modal heads.} We report the mean over three runs. The performance of our method is very similar to the main results when applied to a network for which the uni-modal heads were added after the main training.}
    \label{tab:posthoc}%
    \begin{tabular}{cccccc}
        \toprule
         & \multicolumn{2}{c}{HMDB Near}& \multicolumn{2}{c}{UCF Near} \\
         &  FPR & AUC &  FPR & AUC \\
        \hline
         Original & 33.8\tiny{$\pm$3.0} & 90.3\tiny{$\pm$0.3} & 6.5\tiny{$\pm$1.6} & 98.4\tiny{$\pm$0.3} \\
         Post-hoc & 34.2\tiny{$\pm2.7$} & 90.2\tiny{$\pm$0.1}  & 6.7\tiny{$\pm$1.5} & 98.4\tiny{$\pm$0.1} \\
        \hline
    \end{tabular}%
    % \vspace{-0.4cm}
\end{table}

\section{Applicability beyond action recognition}
We show that our approach can also be used in other domains and tasks by running an experiment on object detection from multi-sensor data. We adapt an RGB \& depth dataset~\cite{lai2011large} for OOD detection by splitting the 50 categories halfway into ID and OOD.
We train a classifier on the ID and show in Tab.~\ref{tab:sensor} that our method can detect OOD samples better than the baselines. We used our default hyperparameters without any tuning.

\begin{table}[t]
    \centering
    % \resizebox{0.8\linewidth}{!}{
    \caption{\textbf{OOD detection for multi-sensor object detection.} Also in this domain, our method outperforms the baselines without hyperparameter tuning.
    }
    \label{tab:sensor}%
    \begin{tabular}{cccccc}
        \toprule
         & DPU\textsubscript{p} & DPU\textsubscript{f} & DPU\textsubscript{h} & Ours \\
        \hline
         AUC & 97.4 & 97.4 & 95.4 & \textbf{98.2} \\
        \hline
    \end{tabular}%
    % }
\end{table}

\section{Additional hyperparameter sensitivity analysis}
To further verify the robustness of our method, we perform the same sensitivity study on a second dataset in Fig.~\ref{fig:sensitivity}, showing equal robustness.

\begin{figure}[t]
  \centering
  \setlength\tabcolsep{1pt}
  \begin{tabular}{ccc}
    \includegraphics[width=0.32\linewidth]{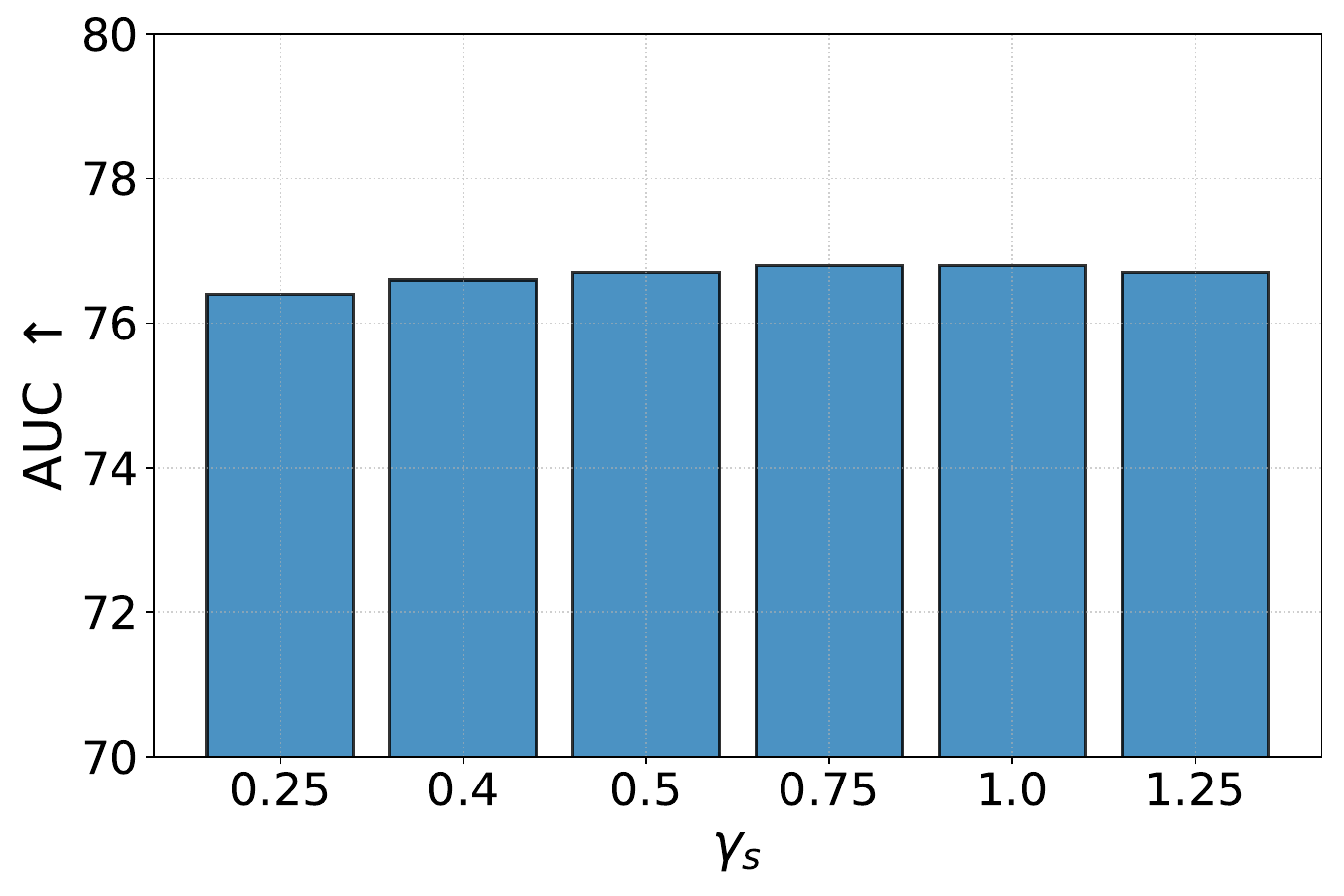} &
    \includegraphics[width=0.32\linewidth]{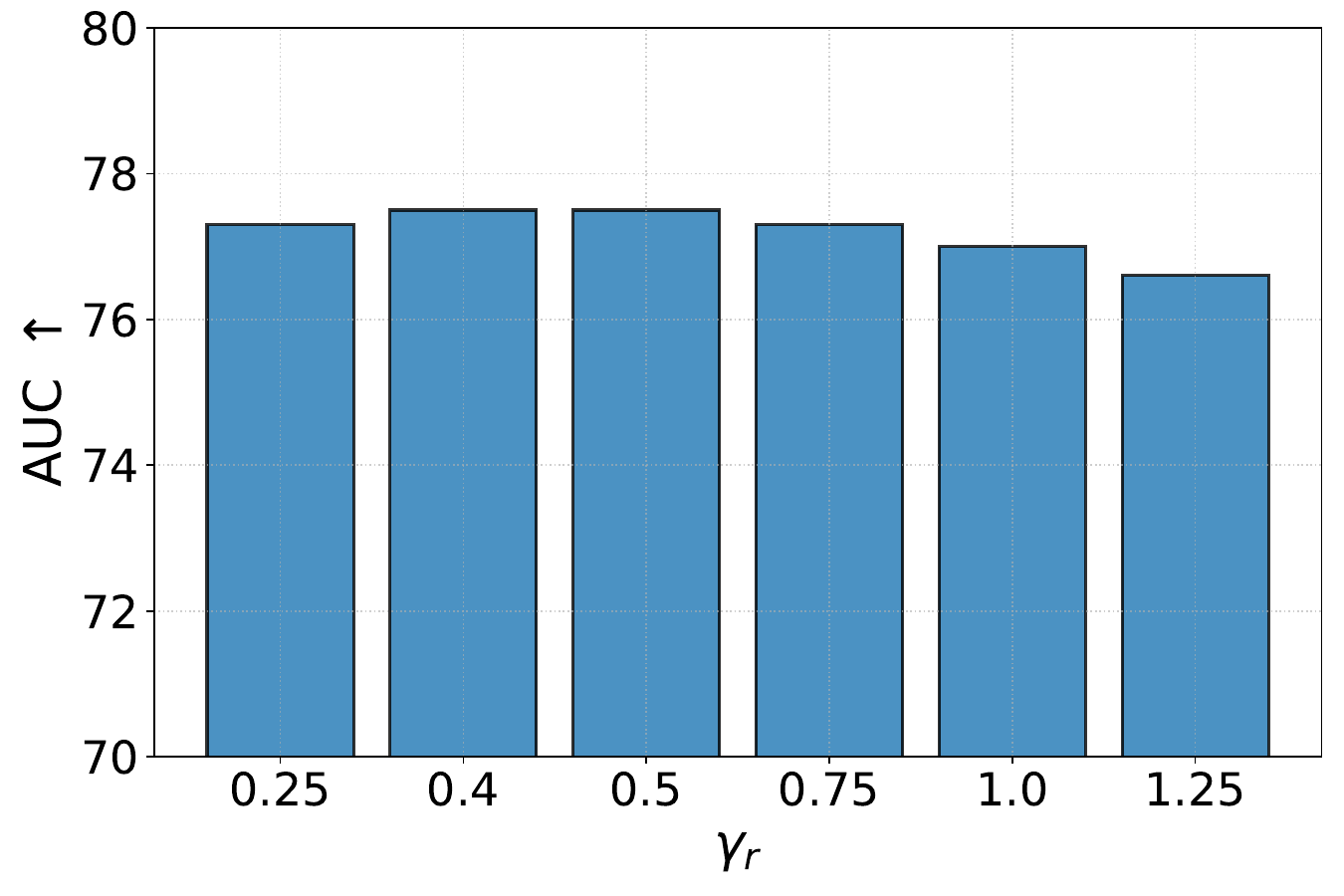} &
    \includegraphics[width=0.32\linewidth]{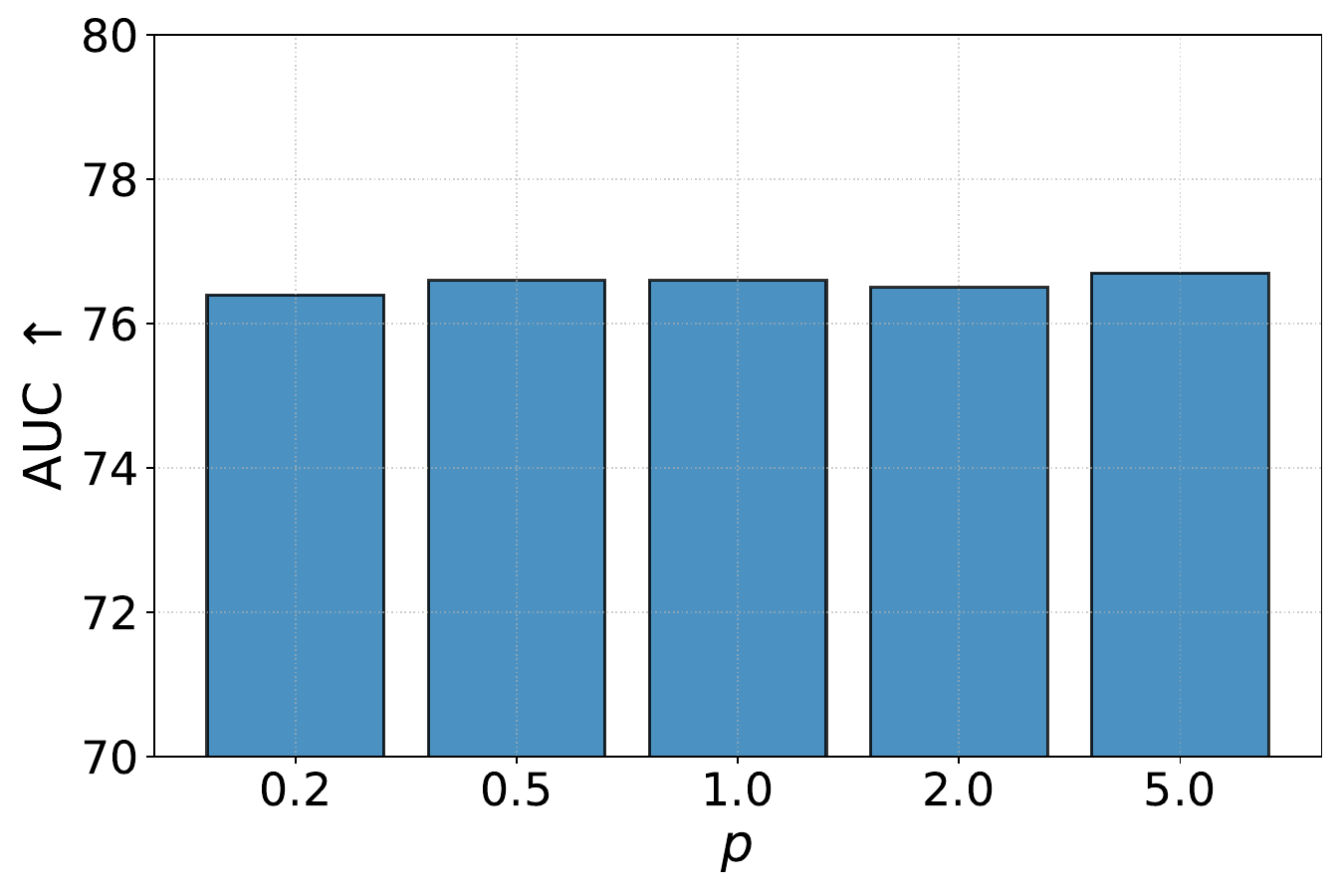} \\
    % (a) & (b) & (c) \\
  \end{tabular}
  \vspace{-5mm}
    \caption{\textbf{Hyperparameter sensitivity} on Kinetics Near-OOD. Our method is robust to a) $\gamma_s$, b) $\gamma_r$, and c) $p$, showing that it does not require task-specific tuning.
    }
    \label{fig:sensitivity}
    \vspace{-3mm}
\end{figure}

\section{Qualitative examples and failure cases}

\begin{figure}[t]
  \centering
  \setlength\tabcolsep{1pt}
  \begin{tabular}{c}
    \includegraphics[width=\linewidth]{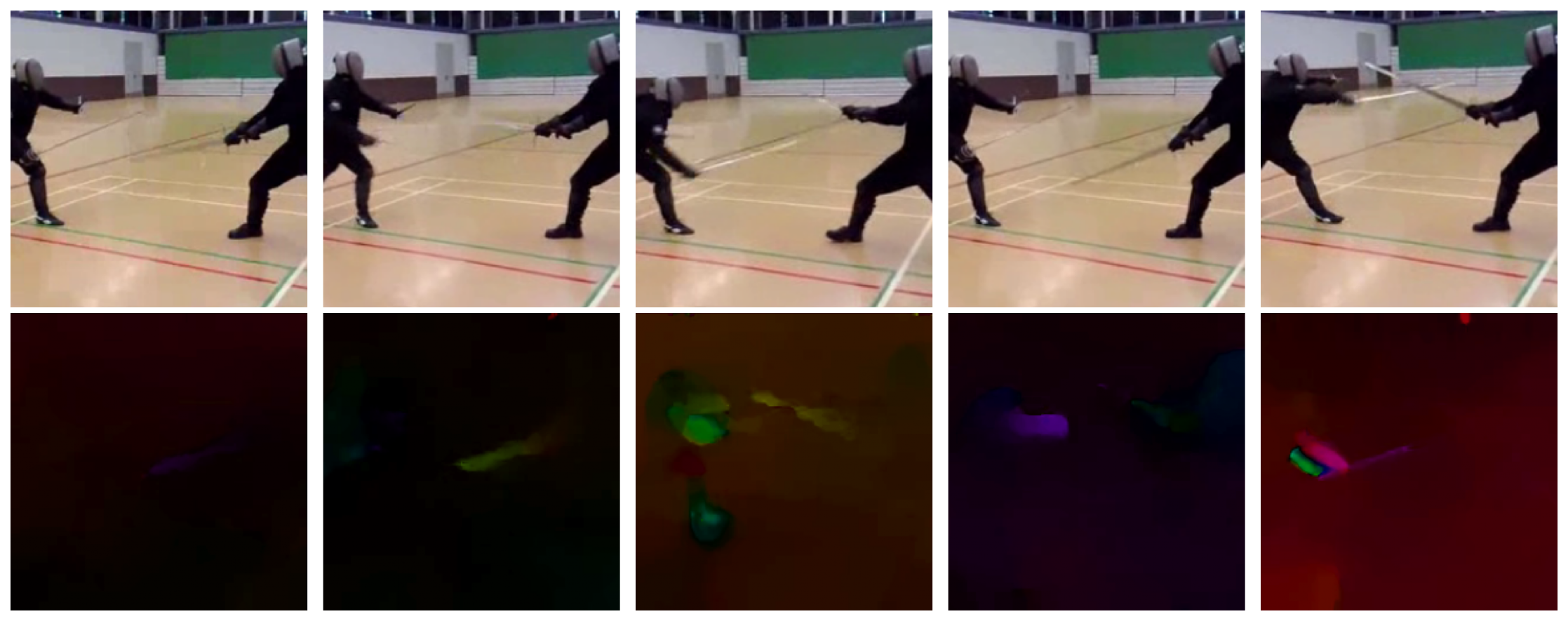}\\
    \textbf{Ours: 95.5 \qquad DPU\textsubscript{f}: 86.2 \qquad DPU\textsubscript{p}: 97.5} \\
  \end{tabular}
    \caption{\textbf{HMDB Near-OOD sample confidently mislabeled as ID by all three methods.} The caption shows the percentage of ID samples that are detected as more OOD for the three methods. We show five frames and the corresponding optical flow of a video from the \textit{sword} class. }
    \label{fig:quali_hmdb_near1}
\end{figure}

\begin{figure}[t]
  \centering
  \setlength\tabcolsep{1pt}
  \begin{tabular}{c}
    \includegraphics[width=\linewidth]{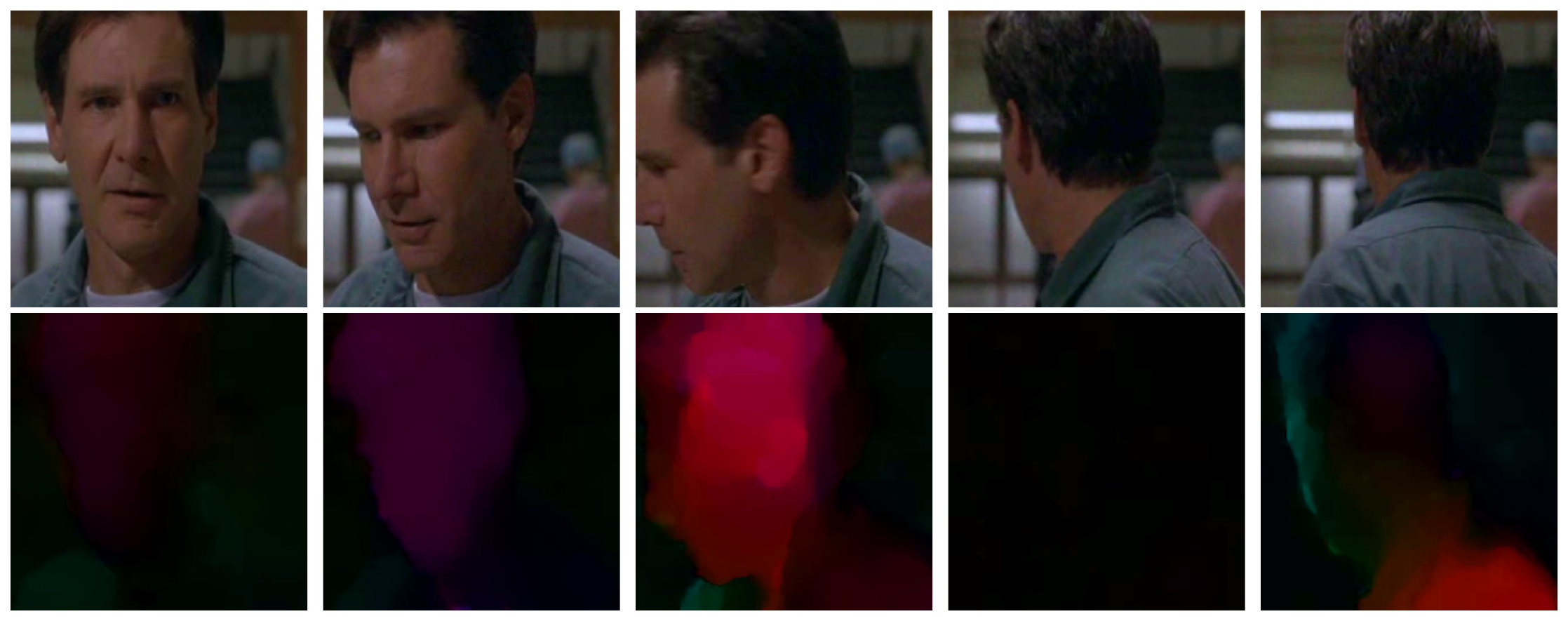}\\
    \textbf{Ours: 5.0 \qquad DPU\textsubscript{f}: 17.2 \qquad DPU\textsubscript{p}: 12.0} \\
  \end{tabular}
    \caption{\textbf{HMDB Near-OOD sample only detected confidently by our method.} The caption shows the percentage of ID samples that are detected as more OOD for the three methods. We show five frames and the corresponding optical flow of a video from the \textit{turn} class.}
    \label{fig:quali_hmdb_near2}
\end{figure}

\begin{figure}[t]
  \centering
  \setlength\tabcolsep{1pt}
  \begin{tabular}{c}
    \includegraphics[width=\linewidth]{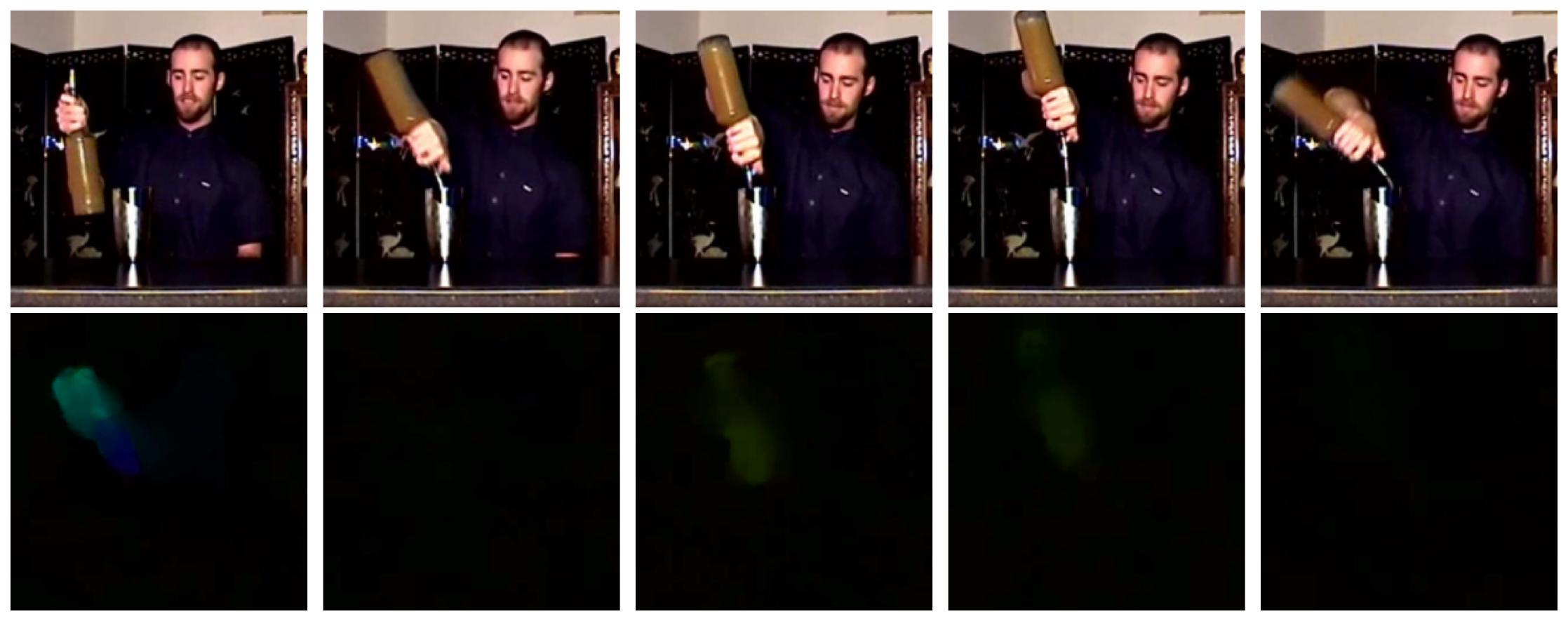}\\
    \textbf{Ours: 8.6 \qquad DPU\textsubscript{f}: 10.6 \qquad DPU\textsubscript{p}: 0.9} \\
  \end{tabular}
    \caption{\textbf{HMDB ID sample detected as mostly OOD by all methods.} The caption shows the percentage of ID samples that are detected as more OOD for the three methods. We show five frames and the corresponding optical flow of a video from the \textit{pour} class.}
    \label{fig:quali_hmdb_near3}
\end{figure}

\begin{figure}[t]
  \centering
  \setlength\tabcolsep{1pt}
  \begin{tabular}{c}
    \includegraphics[width=\linewidth]{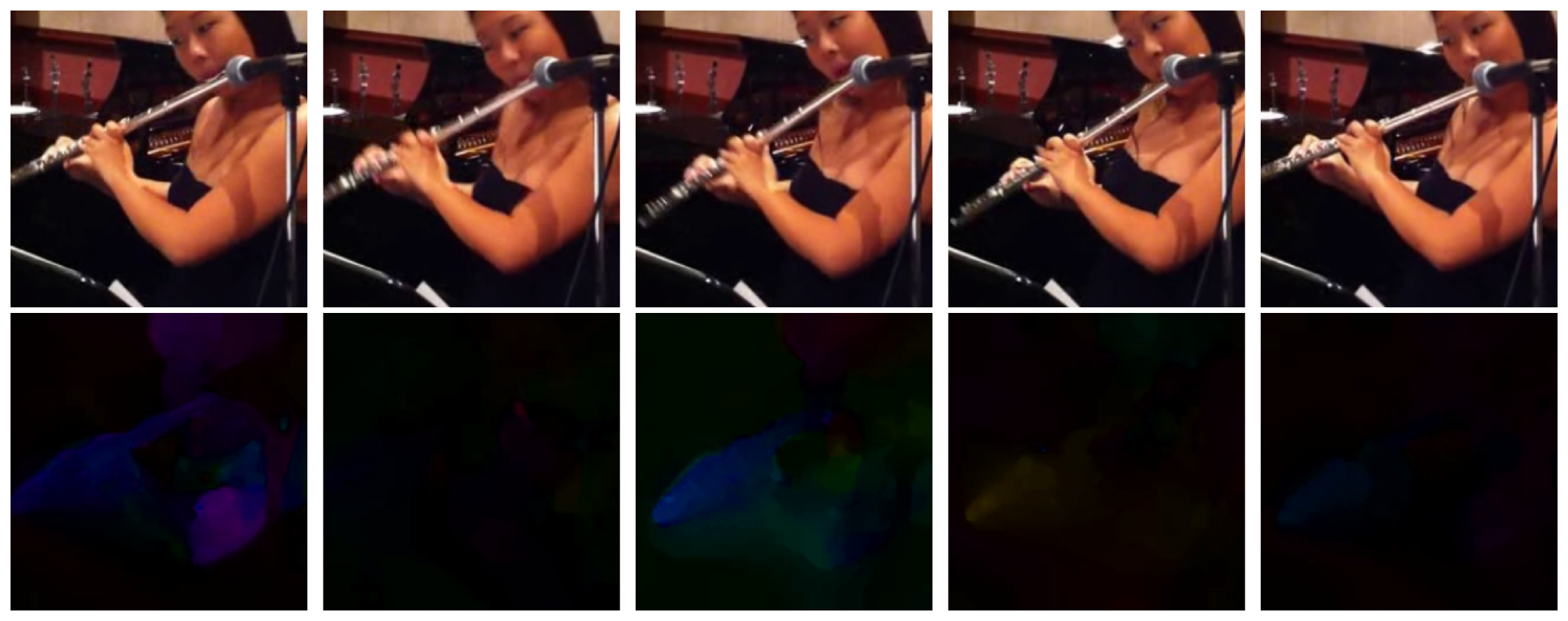}\\
    \textbf{Ours: 0.0 \qquad DPU\textsubscript{f}: 0.5 \qquad DPU\textsubscript{p}: 1.6} \\
  \end{tabular}
    \caption{\textbf{HMDB:UCF Far-OOD sample confidently detected as OOD by all methods.} The caption shows the percentage of ID samples that are detected as more OOD for the three methods. We show five frames and the corresponding optical flow of a video from the \textit{playing flute} class.}
    \label{fig:quali_hmdb_far1}
\end{figure}

We show qualitative examples in Figs.~\ref{fig:quali_hmdb_near1}-\ref{fig:quali_hmdb_far1} along with the OOD score. As different methods have different value ranges, we process them based on the scores of the training set. Specifically, we report the percentage of ID samples that are detected as more OOD. Thus, if a sample has a score of 95.5, this implies 95.5\% of the training samples are seen as more OOD for that method, and the sample is considered very normal. In contrast, a score of 0.0 means that all training samples are seen as more ID and the sample is strongly OOD.

We show the results for three methods: Ours, DPU\textsubscript{p}, and DPU\textsubscript{f}. In general, drawing conclusions from qualitative examples and the corresponding binary scores of OOD detectors is difficult. However, in some cases, there are good reasons why they could be misdetected. For instance, in Fig.~\ref{fig:quali_hmdb_near1}, the OOD ``sword" sample happens in a gym court. ID classes such as ``dribble" can share the same setting, which could lead to spurious correlations. Moreover, the class ``draw sword" in the ID is very similar.
At the same time, our method correctly finds the difficult ``turn" sample in Fig.~\ref{fig:quali_hmdb_near2}, which looks similar to other classes, and does so better than other methods. Some ID samples also score quite high, such as Fig.~\ref{fig:quali_hmdb_near3}, which is a rather unusual ``pour" sample. Finally, Fig.~\ref{fig:quali_hmdb_far1} shows a Far-OOD sample from UCF, which is easily detected by all methods.

\end{document}